\pdfoutput=1

\documentclass[11pt]{article}

\usepackage{ACL2023}
\usepackage{times}
\usepackage{latexsym}

\usepackage[T1]{fontenc}

\usepackage[utf8]{inputenc}

\usepackage{microtype}

\usepackage{inconsolata}
\usepackage[inkscapelatex=false]{svg}
\usepackage{xcolor}
\usepackage{hyperref} 
\usepackage{chngpage}
\usepackage{amsmath}
\usepackage{algorithm}
\usepackage{algorithmic} 
\usepackage{graphicx}
\usepackage{amssymb}
\usepackage{multirow}
\usepackage{diagbox} 
\usepackage{booktabs}
\usepackage{makecell, multirow, booktabs}
\definecolor{airforceblue}{rgb}{0.08, 0.38, 0.74}
\definecolor{blue(ncs)}{rgb}{0.0, 0.53, 0.74}
\hypersetup{
	colorlinks,
	citecolor=airforceblue,
	linkcolor=red,
	urlcolor=blue}

\usepackage{url}            
\usepackage{booktabs}       
\usepackage{amsfonts}       
\usepackage{nicefrac}       
\usepackage{microtype}      
\usepackage[flushleft]{threeparttable}
\usepackage{caption}
\usepackage{subcaption}

\usepackage[inline, shortlabels]{enumitem}
\usepackage{makecell}
\usepackage{listings}
\usepackage{wrapfig,lipsum,booktabs}
\usepackage{tcolorbox}
\usepackage{cleveref}
\usepackage{flushend}
\usepackage{balance}
\usepackage{bbding}
\setlist[itemize]{align=parleft,left=0pt..0.5em}
\setlist[enumerate]{align=parleft,left=0pt..0.5em}
\usepackage{bm}

\title{Fewer is More: Boosting LLM Reasoning with Reinforced Context Pruning}

\author{Xijie Huang$^{1,2*}$    \hspace{5pt} Li Lyna Zhang$^{2\ddagger}$    \hspace{5pt}Kwang-Ting Cheng$^1$   \hspace{5pt} Fan Yang$^2$  \hspace{5pt} Mao Yang$^2$
	\\\fontsize{10}{10} \selectfont{$^1$Hong Kong University of Science and Technology \hspace{9pt} $^2$Microsoft Research \hspace{7pt} }  	
}
\begin{document}
\newcommand{\sysname}{CoT-Influx}
\newcommand{\datasetname}{MRD$^3$-Evol}
\newcommand{\lz}[1]{{\textcolor{red}{\it Lyna: #1}}}	
\newcommand{\xijie}[1]{{\textcolor{blue}{\it Xijie: #1}}}	

\maketitle
\def\thefootnote{$*$}\footnotetext{\fontsize{8.8}{8.8} \selectfont Work was done during the internship at Microsoft Research}
\def\thefootnote{$\ddagger$}\footnotetext{Corresponding author: lzhani@microsoft.com}
\begin{abstract}
Large Language Models (LLMs) have shown impressive capabilities, yet they still struggle with math reasoning. In this work, we propose \textbf{\sysname}, a novel approach that pushes the boundary of few-shot Chain-of-Thoughts (CoT) learning to improve LLM mathematical reasoning. Motivated by the observation that adding more concise CoT examples in the prompt can improve LLM reasoning performance, {\sysname} employs a coarse-to-fine pruner to maximize the input of effective and concise CoT examples. The pruner first  selects as many crucial CoT examples as possible and then prunes unimportant tokens to fit the context window.
A math reasoning dataset with diverse difficulty levels and reasoning steps is used to train the pruner, along with a math-specialized reinforcement learning approach. As a result, by enabling more CoT examples with double the context window size in tokens, {\sysname} significantly outperforms various prompting baselines  across various LLMs (LLaMA2-7B, 13B, 70B) and 5 math datasets, achieving up to 4.55\% absolute improvements. Remarkably, without any fine-tuning, LLaMA2-70B with {\sysname} surpasses GPT-3.5 and a wide range of larger LLMs (PaLM, Minerva 540B, etc.) on the GSM8K. {\sysname} serves as a plug-and-play module for LLMs and is compatible with most existing reasoning prompting techniques, such as self-consistency and self-verification.

\end{abstract}

\section{Introduction}

Large Language Models (LLMs) have demonstrated remarkable capabilities across a range of tasks~\cite{gpt3,gpt4}. However, it remains a significant challenge to improve LLM performance on reasoning tasks, especially for smaller LLMs like LLaMA~\cite{llama} on math reasoning.

While existing efforts focus on optimizing Chain-of-Thought (CoT) prompts~\cite{cot, selfconsistency,treeofthought} and fine-tuning LLMs~\cite{luo2023wizardmath} under the zero-shot setting, the potential of few-shot learning in improving LLM reasoning has not been fully explored. Inspired by the human reasoning process, we propose the hypothesis: \textit{if LLMs are exposed to more step-by-step problem-solving examples (i.e., CoTs) before answering questions, it could potentially improve LLMs reasoning capability to generate a correct solution}.  This leads to our question: \textit{what's the boundary of LLM reasoning capability achievable through inputting more CoT examples?}

However, we face two major obstacles. 
First, the limited token length of LLMs' context window restricts the number of few-shot examples. 
Extending the context window is one solution, but it requires expensive fine-tuning and increases inference overhead~\cite{pi,peng2023yarn}.
While prompt compression~\cite{li2023compressing,llmlingua} is another approach, it underperforms in math reasoning. Tokens like numerical and format ones, though identified redundant, are crucial for few-shot math problem solving.

Second,  {it's challenging to select helpful CoT examples}. 
Section~\ref{sec:pilot_study} reveals that random choices can even harm reasoning performance. Existing retrieval-based methods~\cite{topk,scarlatos2023reticl} are not tailored  for math reasoning, making them suboptimal. These retrieved examples are model-agnostic, while we found that different LLMs  favor CoT examples of varying characteristics (e.g., diverse difficulty levels).

In this work, we propose \textbf{\sysname}, which addresses all the above challenges and pushes the boundaries of utilizing few-shot learning to improve  LLM math reasoning capability. {\sysname} is motivated by the observation that \textit{current LLM context window has not been fully utilized due to redundancy at both the example and token levels in natural language input}. As such, these redundant inputs can be pruned to free up space for more informative context. The central idea of {\sysname} is to input long lengthy CoT examples, select the crucial examples for the target LLM, and then prune redundant tokens to fit within the original LLM context window. As a result, by inputting much more helpful CoT examples, each composed solely of informative tokens and with a shorter length, we greatly improve LLM ability to solve math problems. Moreover, as all these inputs remain within the context window, we do not increase any inference overhead. This stands in stark contrast to other methods~\cite{hao2022structured,pi}.

{\sysname} treats the target LLM as a black box, and serves as a plug-and-play module for LLMs as shown in Fig.~\ref{fig:overview}. The key module is a coarse-to-fine pruner involving two steps: (i) a shot pruner first selects the most helpful CoT examples from a large batch of shots,  and (ii) a token pruner then removes unimportant tokens from these selected CoT examples. To effectively train the pruner module tailored for math reasoning, {\sysname} is built upon the following novel techniques. 

First,  {\sysname} requires a CoT dataset for training and inference. Existing CoT examples,  heavily reliant on costly human engineering, often struggle with diversity and quality.  To address this, we employ GPT-4~\cite{gpt4} and Evol-Instruct~\cite{xu2023wizardlm} to create a math reasoning dataset, called \textit{MRD$^3$}.
With problems of varying difficulty and reasoning steps, MRD$^3$ enables  {\sysname} to generalize across a wide range of math problems.

Second, training the pruner presents two challenges: (1) since we identify discrete tokens before the LLM tokenizer, the loss gradient cannot be backpropagated through the tokenizer to update the pruner; (2) The high difficulty of many math problems, which consistently yield incorrect answers regardless of the quality of compressed few-shot examples, poses a challenge to the effective training of the pruner.  
To this end, we introduce a novel training approach with reinforcement learning to mitigate the gradient issue. 
We design a reward function to measure the LLM loss, few-shot math reasoning effectiveness, and token length constraints. Then, we design a difficulty-aware dataloader filtering appropriate problems and introduce two techniques to stabilize the RL training.

Extensive experiments on various LLMs and five
math datasets demonstrate the effectiveness of {\sysname}. {\sysname} significantly boosts LLM reasoning capability,  achieving 1.36\%-14.09\% absolute improvements over SOTA baselines, and establishes a new prompting-based benchmark in math reasoning accuracy without any fine-tuning or additional inference costs. Remarkably, LLaMA2-70B with {\sysname} outperforms a broad range of larger LLMs and surpasses GPT-3.5 by 2.5\% on GSM8K.
Moreover, {\sysname}  excels over retrieval and prompt compression baselines in example selection and identifying crucial tokens.

\section{Related Works}
\vspace{-1ex}
\noindent\textbf{LLMs for Math Reasoning}. Drawing from the Chain-of-Thought (CoT)~\cite{cot}, recent research has greatly improved the reasoning capabilities of LLMs by providing step-by-step reasoning paths. The main efforts are twofold: enhancing CoT prompts, such as Program-of-Thoughts~\cite{pot}, Tree-of-Thoughts~\cite{treeofthought}, and Everything-of-Thoughts~\cite{ding2023everything}, and innovating CoT-based training data for fine-tuning LLMs like WizardMath~\cite{luo2023wizardmath}. 

However, most works focus on the zero-shot setting with only task instruction or CoT prompts,  leaving the potential of few-shot CoT largely untapped. We explore leveraging few-shot CoT learning to improve LLMs' math reasoning capabilities.

\noindent\textbf{Prompt Compression}. To address the challenge of limited few-shot examples due to restricted context window length, one related work involves prompt compression.  Key approaches  include:  (1) token pruning~\cite{ltp,top}; (2) soft prompt compression methods~\cite{wingate2022prompt,mu2023learning,chevalier2023adapting,ge2023context}; and (3) information-entropy-based approaches~\cite{li2023compressing,llmlingua}.

 However, they do not effectively solve our problem for two reasons. First, they prune tokens based on designed metrics, often failing to remove  redundancy of the entire CoT examples. 
Second, some tokens such as numerical and format tokens, although redundant, are crucial for math reasoning.

\noindent\textbf{Prompt Retrieval} optimizes task performance by selecting high-quality few-shot examples using either heuristics or a supervised retriever model. Heuristic methods, such as the widely used TopK retrieval~\cite{topk, gao2021making}, BM25~\cite{robertson2009probabilistic}, VoteK~\cite{hongjin2022selective}, and entropy~\cite{lu2022fantastically}, select examples based on semantic similarity. Recently, supervised-based methods like  EPR~\cite{rubin2021learning}, LLM-R~\cite{wang2023learning}, and IDS~\cite{qin2023context} have been proposed, which train a retrieval model to learn better example selection.

However, these methods are sub-optimal for math reasoning, as they retrieve model-agnostic examples. In contrast, LLMs with different capabilities favor CoT examples of varying complexities. Moreover, they don't account for token redundancy, which restricts the number of retrieved examples.

\vspace{-1ex}
\section{Pilot Study}\label{sec:pilot_study}
\vspace{-1ex}
\begin{figure}[t]
	\centering
	\includegraphics[width=1\columnwidth]{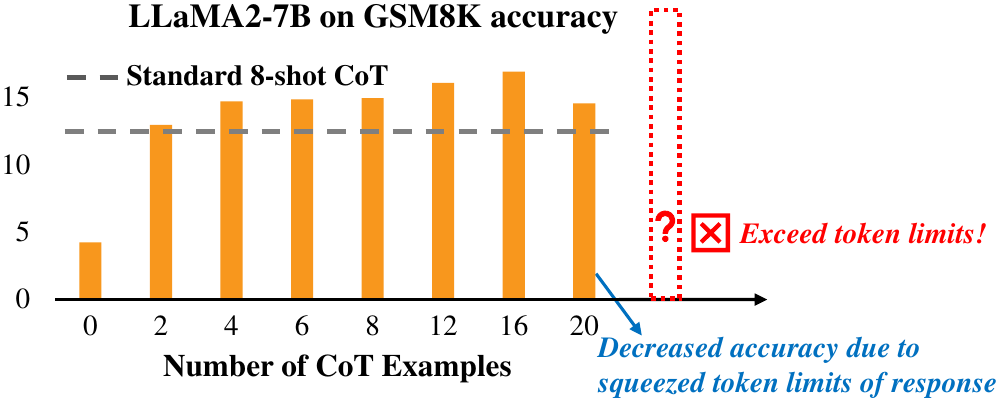}
	\vspace{-4ex}
	\caption{LLaMA2-7B reasoning accuracy under an increasing number of TopK retrieved CoT examples.}
	\label{fig:cotnumber}
 \vspace{-2ex}
\end{figure}
This section presents our key observations of few-shot learning in improving LLMs math reasoning, upon which the {\sysname} design is based. Note that experiments are done with our proposed CoT dataset, {MRD$^3$}, as introduced in Sec.~\ref{sec:dataset}.

\noindent\textbf{Observation 1}: \textit{LLMs can improve reasoning with more helpful CoT examples, but the current context window restricts the number of  CoT examples}.

A standard practice for evaluating LLMs' math reasoning capability is the use of 8-shot manually-designed CoTs~\cite{cot}. We increase the number of CoT shots to see if reasoning accuracy improves. To avoid poor-quality examples, we use the TopK method~\cite{topk} to select the $k$ most relevant CoT examples for each question. Given LLaMA2's context window limit of 4096 tokens, we could only input up to 20 CoT examples\footnote{The input token length is less than the context window token limit, as the answer generation also shares this limit. }. As Fig.~\ref{fig:cotnumber} shows, increasing CoT examples improves LLaMA2-7B's reasoning accuracy on the GSM8K dataset, significantly outperforming the standard 8-shot setting.  However, the limited LLM context window hinders the full potential of few-shot CoT learning for improving math reasoning. For instance, even with 20 CoTs not hitting the token limit, accuracy drops as the large input context limits the LLM's response space.

\noindent\textbf{Observation 2}: \textit{CoT example selection is crucial for math reasoning. Simply adding  CoT examples randomly doesn't boost performance}.

The prior study suggests that more CoT examples can improve LLM reasoning performance. However, the quality of CoT examples is crucial to the final performance. As shown in Table~\ref{tbl:cotquality}, even with up to 16 CoT shots, random selection underperforms the standard 8-shot setting, which is manually curated for quality. 

\begin{table}[ht]
\vspace{-1ex}
	\caption{The selection of CoT examples heavily impacts LLM reasoning performance.}
	\label{tbl:cotquality}
	\vspace{-2ex}
	\small
	\centering
	\begin{tabular}{@{\hskip5pt}c@{\hskip7pt}c@{\hskip7pt}c@{\hskip7pt}c@{\hskip5pt}}
		\toprule
		\multirow{2}{*}{Model}&	\multirow{2}{*}{Manual 8 Shots}&\multicolumn{2}{c}{Method (16 Shots)}\\
		\cmidrule{3-4}
		& & Random 1 & Random 2\\
		\midrule
		LLaMA2-7B& 13.79&12.36&13.27\\
		LLaMA2-13B&27.82&23.04&23.28\\
		\bottomrule
	\end{tabular}
	\vspace{-1ex}
\end{table}

\noindent\textbf{Observation 3}: \textit{A CoT example contains redundant tokens for math reasoning, which can be pruned to free up space for more informative content.} 

\begin{figure}[ht]
	\centering
	\includegraphics[width=1\columnwidth]{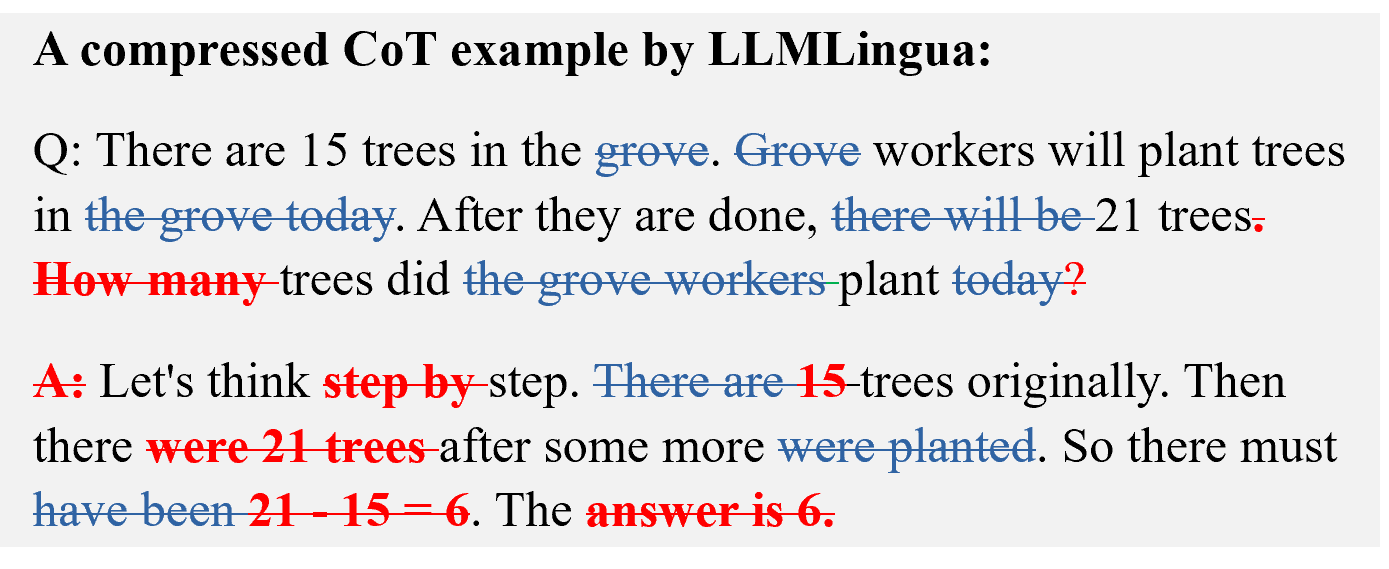}
	\vspace{-4ex}
	\caption{A compressed CoT example using the prompt compression tool of LLMLingua~\cite{llmlingua}. The pruned tokens contain truly redundant tokens (colored in blue) and crucial tokens (colored in red). }
	\label{fig:compressed_example}
	\vspace{-2ex}
\end{figure}

Observation 2 indicates that few-shot CoT examples contain useless or even harmful examples that can be pruned. We further observe that a CoT example often has redundant tokens. For instance, the blue tokens in Fig.~\ref{fig:compressed_example} can be removed without affecting LLM performance. However, identifying redundant tokens for math reasoning poses a challenge. Simply using existing prompt compression methods~\cite{llmlingua,li2023compressing} leads to a significant performance decline. Fig.~\ref{fig:compressed_example} shows a compressed example using LLMLingua~\cite{llmlingua}. Some numerical and format tokens (colored in red), while identified as redundant, are crucial for LLM to comprehend the context for solving a math problem.

\vspace{-1ex}

\begin{figure*}[ht]
	\centering
	\includegraphics[width=1\textwidth]{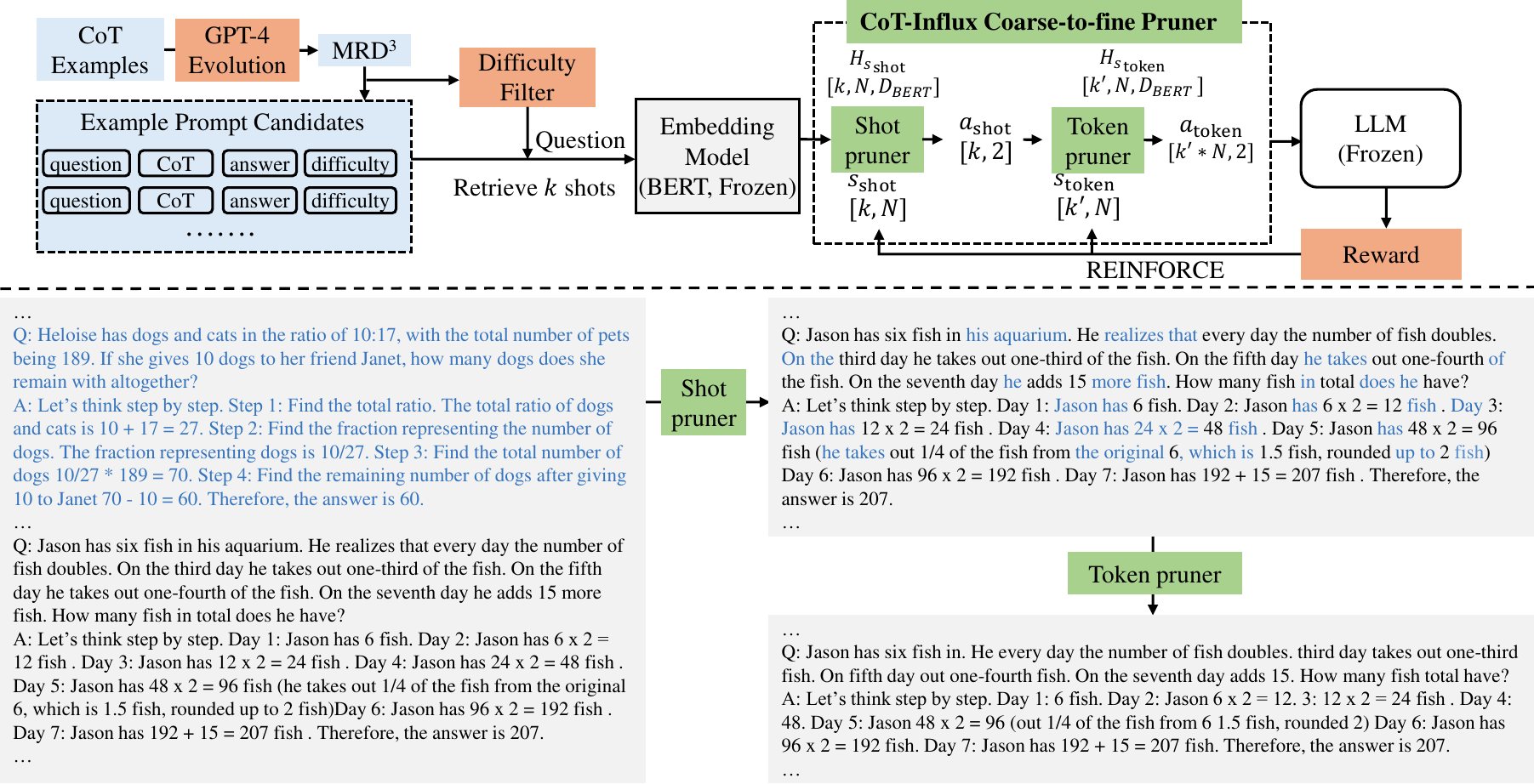}
	\vspace{-4ex}
	\caption{\textit{Above}: The overview procedure of {\sysname}; \textit{Below}: an example illustrating the use of {\sysname} to first prune entire CoT examples and then prune tokens. }
	\label{fig:overview}
	\vspace{-2ex}
\end{figure*}

\vspace{-0.5ex}
\section{{\sysname} Methodology}
\vspace{-0.5ex}
Motivated by our observations, this section introduces {\sysname}, which maximizes CoT examples within the LLM context window by identifying the most important CoT examples and tokens from long lengthy input contexts. 

\subsection{CoT Dataset Collection}
\vspace{-0.5ex}
\label{sec:dataset}
We start by collecting a high-quality math reasoning dataset, comprising diverse CoT examples with varying steps and difficulties. We merge the training set of GSM8K~\cite{gsm8k}, MAWPS, MAWPS-single~\cite{mawps}, and 1000 random examples from AQuA~\cite{ling2017program} to create an initial dataset of 9.7K question-answer pairs. Then, we prompt GPT-4 to generate formatted CoT reasoning steps. Notably, it's crucial to maintain a consistent format for each example in few-shot learning.  Our dataset also assigns a difficulty score from 1 to 10 for each question, based on GPT-4's evaluation, where 1 signifies the easiest questions and 10 is the most difficult. 

We observe that most questions in this initial dataset score between 2-4. To improve difficulty diversity, we use GPT-4 to mutate questions, generating corresponding CoTs with varied difficulty levels. We apply 5 mutation schemes, three to increase reasoning difficulty and two to simple questions. The final dataset is referred to as \textit{M}ath \textit{R}easoning \textit{D}ataset with \textit{D}iverse \textit{D}ifficulty (\textit{MRD$^3$}).

\subsection{Problem Formulation}
\label{sec:problem_formulation}
Let $\mathcal{D}$ denote the  CoT dataset (i.e., the MRD$^3$), and $\hat{\mathcal{D}}=\{x^\text{cot}_i\}_{i=1}^k$ be a subset of $k$ CoT examples, each composed of a question, reasoning steps, and an answer. 
The total number of tokens in these $k$ CoT examples far exceeds the LLM context window length limit of $T$.    {\sysname} is designed to perform a two-step pruning process:
\vspace{-1ex}
\begin{equation}
\resizebox{0.43\textwidth}{!}{$
\hat{\mathcal{D}}=\{x^{\text{cot}}_i\}_{i=1}^{k} \xrightarrow{\text{Shot Pruner}} \{ x^{\text{cot}}_j\}_{j=1}^{k'} \xrightarrow{\text{Token Pruner}} \{ \hat{x}^{\text{cot}}_{j}\}_{j=1}^{k'}  $}
\end{equation}
Initially, non-useful CoT examples are pruned from $\hat{\mathcal{D}}$, resulting in a reduced set of $k'$ examples. Then, for each retained CoT example $x^\text{cot}$, redundant tokens are pruned, yielding a shorter example,  $\hat{x^\text{cot}}$.

Let $x^\text{question}$ denote the question that LLM is tasked to solve. For final input $x^\text{input}$, we concatenate all tokens from $\{ \hat{x}^{\text{cot}}_{j}\}_{j=1}^{k'}$ and place them before $x^\text{question}$. 
Our goal is to optimize the input $x^\text{input}$, so that LLM can correctly answer the question under $x^\text{input}$.
Meanwhile, the token count of $x^\text{input}$,  $t$, must be less than the LLM context window limit $T$. Formally, we optimize the following: 
\vspace{-0.5ex}
 \begin{equation} \label{eq:target}
 	\resizebox{0.43\textwidth}{!}{$
 	\begin{aligned}
 			\min_{ \hat{\mathcal{D}}\subseteq \mathcal{D}} & L_{\text{LLM}}\left(x^{\text{input}}\right), \,
 			\max_{ \hat{\mathcal{D}}\subseteq \mathcal{D}} R_{\text{Acc}}\left(y_{\text{LLM}}\left(x^{\text{input}}\right), y^{\text{answer}}\right),
 			\\ s.t. \quad  &t\left(x^\text{input}\right)=\sum_{1}^{k'} |\hat{x}^{\text{cot}}|+|x^\text{question}|\leq T
 		\end{aligned}
 $}\vspace{-0.5ex}
 \end{equation}
where $L_\text{LLM}$ is LLM  loss, and $R_\text{Acc}$ evaluates if LLM's answer $y_{\text{LLM}}(x^{\text{input}})$ matches the correct answer $y^\text{answer}$, this will be elaborated in Sec.~\ref{sec:training}.

\noindent\textbf{Overview}. Fig.~\ref{fig:overview} illustrates our approach. The core component is a lightweight,  plug-and-play module (Sec.~\ref{sec:prunerdesign}),  which consists of a small text embedding extractor and a coarse-to-fine pruner. 

To train the pruner, we face the challenge of gradient backpropagation when pruning discrete tokens outside the LLM. The  LLM loss gradient cannot be backpropagated through the tokenizer. To address this, 
we design a multi-objective reward function and use reinforcement learning for effective training (Sec.~\ref{sec:training}). The overall training process is outlined in Algorithm~\ref{alg:cot-max}.

\subsection{Coarse-to-fine Pruner Design}
\label{sec:prunerdesign}

\noindent\textbf{Text embedding extractor}. As {\sysname} serves as an external module, we need to extract text embedding as prediction features. However, it's non-trivial to extract features for long inputs beyond the LLM context window. To address this, 
we use a small encoder model, BERT-Large~\cite{bert},  to extract sentence-level (i.e., a CoT example) embedding instead of extracting token embedding from the entire long context. For a batch of $k$ CoT examples, each example is padded to $N$=512 tokens. BERT then inferences these examples to obtain the final layer of text embedding, denoted as $H_{s_\text{shot}}\in\mathbb{R}^{k\times N\times D_{BERT}}$, where $D_{BERT}$ is BERT's hidden dimension size.

\noindent\textbf{State}. As shown in Fig.~\ref{fig:overview}, we define state $s_\text{shot}\in\mathbb{R}^{k\times N}$ for the first shot pruner, representing the input batch of $k$ CoT examples $\in\hat{\mathcal{D}}$. For the second token pruner, we define state $s_\text{token}\in\mathbb{R}^{k'\times N}$, which represents all remaining tokens after the shot pruner. $k'$ is the number of retained examples.

\noindent\textbf{Action}. Let $a_\text{shot}$ and $a_\text{token}$ denote the actions predicted by the shot and token pruner, respectively. The action space is defined as  \{0, 1\}, where 1 signifies retention and 0 indicates pruning.  $a_\text{shot}$ determines whether each CoT example should be pruned, while $a_\text{token}$ predicts the pruning of each token in the retained CoT examples.

\noindent\textbf{Two-stage policy network}. The pruner module is a two-stage policy network, each stage is a two-layer feed-forward network (MLP) with GELU activation. This module outputs a continuous categorical distribution $\pi$,  used for sampling discrete actions (i.e., \{0, 1\}).  Let $\theta$ denote the MLP's trainable parameters and $\sigma(\cdot)$ the sigmoid function. Based on the current states $\{s_\text{shot}, s_\text{shot}\}$ and the hidden states $\{H_{s_\text{shot}}, H_{s_\text{token}}\}$, the policy network sequentially make two action predictions as follows:
\begin{equation} \label{eq:shot_policy}
	\small
\displaystyle \pi(a_\text{shot}|s_\text{shot};\theta) = \sigma\left( \text{MLP} \left( H_{s_\text{shot}}\right)\right)
\end{equation}
\vspace{-2ex}
\begin{equation} \label{eq:token_policy}
	\small 
	\displaystyle
\pi(a_\text{token}|s_\text{token};\theta) = \sigma\left(\text{MLP}\left(H_{s_\text{token}}\right)\right),
\end{equation}
where $a_\text{shot}$ and $a_\text{token}$ are the predicted actions, sequentially predicting whether to prune each of the $k$ CoT examples and each token within the retained examples, respectively. 
We predict the discrete action by sampling from the categorical distribution $\pi$ with a softmax function.

\begin{algorithm}[t]
	\small
	\caption{Pruner Training and Inference}
	\label{alg:cot-max}
	\textbf{Input:} target LLM, dataset $\mathcal{D}$,   number of CoTs $k$, token limit $T$, manual few-shot cot $x^{\text{few-shot}}$, repeat $t_{\text{repeat}}$ \\
	\vspace{-2.5ex}
	\begin{algorithmic}[1]
		
		\STATE $\blacktriangleright$ \textbf{Phase 1: MRD$^3$ preperation}
		\STATE Perform evolution and difficulty evaluation to get $\mathcal{D}$;
		\STATE Use the difficulty filter and split $\mathcal{D}$ into questions set $\mathcal{D}_{\text{question}}$ and CoT candidates set $\mathcal{D}_{\text{cot}}$
		\STATE $\blacktriangleright$ \textbf{Phase 2: Training the two-stage pruner (1 epoch)}
		
		\FOR{$(x^{\text{question}},  y^{\text{answer}})$ in $\mathcal{D}_{\text{question}}$}
		
		\STATE Retrieve Top-$k$ examples $\hat{\mathcal{D}}=\{x^{\text{cot}}\}^k_{i=1}$ from $\mathcal{D}_{\text{cot}}$
		
		\STATE $H_{s_{\text{shot}}}$ = BERT$\left(\{x^{\text{cot}}\}^k_{i=1}\right)$ 
		
		\FOR{$j$=$1$ to $t_{\text{repeat}}$}
		
		\STATE Get $\pi\left(a_\text{shot}|s_\text{shot};\theta\right)$ with 
		Eq.~\ref{eq:shot_policy}, sample $a_\text{shot}$
		
		\STATE $\{x^{\text{cot}}\}^k_{i=1} \xrightarrow{}\{x^{\text{cot}}\}^{k'}_{i=1}$
		\STATE $H_{s_{\text{token}}}$ = BERT$\left(\{x^{\text{cot}}\}^{k'}_{i=1}\right)$
		
		\STATE Get $\pi\left(a_\text{token}|s_\text{token};\theta\right)$ with 
		Eq.~\ref{eq:token_policy}, sample $a_\text{token}$
		
		\STATE $\{x^{\text{cot}}\}^{k'}_{i=1}\xrightarrow{}\{\hat x^{\text{cot}}\}^{k'}_{i=1}$
		
		\STATE  $x^{\text{input}} = \left(\{\hat x^{\text{cot}}\}^{k'}_{i=1}, x^{\text{few-shot}}, x^{\text{question}}\right)$
		
		\STATE Output ${\text{LLM}}({x}^{\text{input}})$; Compute $R$ with Eq.~\ref{eq:total_reward}
		\ENDFOR
		
		\STATE Compute policy gradient using Eq.~\ref{eq:gradient}, update $\theta$ 
		\ENDFOR 
		
		\STATE $\blacktriangleright$ \textbf{Phase 3: LLM reasoning with pruner and MRD$^3$}
		
		\STATE Retrieve Top-$k$ shots $\{x^{\text{cot}}_{q}\}^k\in\mathcal{D}$ for  target question $q$
		
		\STATE Do pruning: $\{x^{\text{cot}}_{ q}\}^k\xrightarrow{\theta}\{\hat x^{\text{cot}}_{ q}\}^{k'}$, reconstruct $\{\hat x^{\text{cot}}_{q}\}^{k'}$ 
		
		\STATE $x^{\text{input}}_{q} = \left(\{\hat x^{\text{cot}}_{q}\}^{k'}, x^{\text{few-shot}}, x^\text{question}_q\right)$
		
		\STATE Get LLM reasoning output $ {\text{LLM}}({x}^{\text{input}}_{ q})$
	\end{algorithmic}
\end{algorithm}

\subsection{End-to-end RL Optimization}
\label{sec:training}
\noindent\textbf{Multi-objective Reward}. 
Our objective in Eq.~\ref{eq:target} is to train the pruner module to identify the most crucial CoT examples and useful tokens for math problem solving, while keeping the final tokens within the original LLM context window. 
To achieve this, we design a multi-objective reward.

Let $x^\text{input}$ be the final input to  LLM, which includes the retained CoT tokens from the policy network and the target question. $t$ represents the token count of $x^\text{input}$, and $T$ is the token count limit. The reward $R$ is defined as follows:
\vspace{-1ex}
\begin{equation} \label{eq:total_reward}
	\small
	\displaystyle	R\left( x^\text{input}\right) = (\frac{1}{1+L_\text{LLM}}+R_{\text{Acc}}) \times \left[\frac{t}{T}\right]^w
\end{equation}
where the first term evaluates the effectiveness of inputted CoT tokens, and the second term ensures they are within the LLM context window. $L_\text{LLM}$ is LLM's prediction loss under $x^\text{input}$, $R_\text{Acc}$ evaluates the reasoning accuracy (to be discussed later). $w$ is a hyperparameter that penalizes the token count $t$ for being too short (i.e., $w<0$) or exceeding (i.e.,$w>0$ ) the token limit $T$. 

In addition to $L_\text{LLM}$, we introduce $R_\text{Acc}$ to evaluate math reasoning accuracy. This is because $L_\text{LLM}$,  the average loss of generated tokens, doesn't fully reflect LLM's ability to generate correct answers. Specifically, $R_\text{Acc}$ is set to 1 for a correct answer and 0 for an incorrect one. Notably, we found that if the format or crucial tokens are pruned, LLM struggles to interpret the input context correctly, leading to irrelevant answers for math problem solving.   In such cases, we penalize $R_\text{Acc}$ with a value of -0.1.

\noindent\textbf{Optimization with REINFORCE}. We employ reinforcement learning to maximize the reward and train the two-stage policy network. According to REINFORCE~\cite{reinforce}, the network parameters are updated by the gradients: 
\vspace{-1ex} 
\begin{equation} \label{eq:gradient}
	\small
	\resizebox{0.35\textwidth}{!}{$ R \cdot \nabla_\theta \text{log} \pi(a_\text{shot}|s_\text{shot})\pi(a_\text{token}|s_\text{token})$}\vspace{-1ex} 
\end{equation}

Notably, as shown in Fig.~\ref{fig:overview}, only the parameters of the policy network require training. The embedding extractor and LLM are frozen, thus, the overall training overhead is lightweight.

\noindent\textbf{Difficulty-aware data filter}.
Existing LLMs, particularly smaller ones, underperform in math reasoning. If the question is too challenging for LLMs, the answer will always be incorrect, regardless of the quality of compressed few-shot CoT examples, making it challenging to effectively train our pruner module. To address it,  
we use a difficulty filter to sample a math question set $\mathcal{D}_{\text{question}}$ from $\mathcal{D}$, which includes only easy questions with a difficulty score below a threshold $d_{\text{thr}}$.  During training, each question in $\mathcal{D}_{\text{question}}$ samples a batch of $k$ CoT examples from $\mathcal{D}_{\text{cot}}$, where $\mathcal{D}_{\text{cot}}$ is the CoT candidate set sampled from $\mathcal{D}$.

\noindent\textbf{Stabilize the training}. Another challenge is that pruning CoT and tokens during training introduces instability, making it difficult for effective training.

First, despite the optimization of  question set $\mathcal{D}_{\text{question}}$ through the filter, 
LLM performance for a randomly sampled question under different few-shot prompts can still be unpredictable. This unpredictability, where a question might yield correct results under low-quality pruned prompts and a complex question might fail under carefully pruned prompts, can affect the pruner's training effectiveness. To address this, we continuously repeat a sampled question multiple times, $t_\text{repeat}$, each time with a different pruned few-shot prompt from the pruner.  Moreover, we use exponential moving average (EMA) to smooth reward $R_\text{Acc}$ in Eq.~\ref{eq:total_reward}.

Second, during the early training, our pruner module makes random decisions, leading to arbitrary removal of CoT examples and tokens. These randomly pruned few-shot prompts can cause instability in RL training. Empirically, we append the manually-designed 8-shot CoTs~\cite{cot} to the pruned prompts.  
This ensures a good lower bound and stabilizes the training.

\begin{figure}[t]
	\vspace{-1ex}
	\centering
	\includegraphics[width=0.48\textwidth]{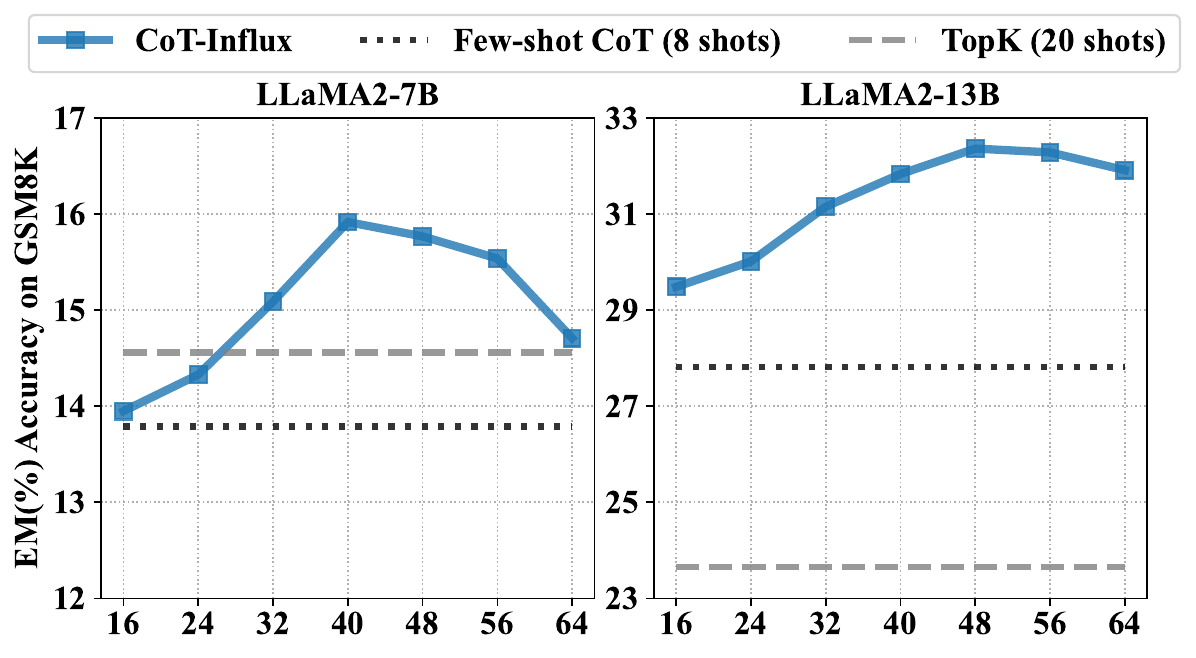}
	\vspace{-4ex}
	\caption{EM(\%) accuracy on GSM8K with inputting different number of  CoT examples for {\sysname}.}\label{fig:shot_number}
	\vspace{-2ex}
\end{figure}
\begin{table*}[t]
	\caption{Comparison  of EM (\%) accuracy on GSM8K with state-of-the-art baselines. Note that the 20 CoT shots of retrieval baselines are the max number, given that the context window limit of LLaMA2 is 4096 tokens.}
	\label{tab:maintable}
	\vspace{-2ex}
	\centering
	\resizebox{\textwidth}{!}{
		\begin{tabular}{lcccccccc}
			\toprule
			Method   & \#Input CoT shots & \#Average tokens & LLaMA2-7B &  LLaMA2-13B &  LLaMA2-70B  \\
			\midrule
			Zero-shot             &  0 & - & 4.25 &  5.84& 11.45\\
			Zero-shot-CoT~\cite{kojima2022large}   &  0 & - & 1.74 & 12.28& 21.91\\
			Few-shot-CoT~\cite{cot}   & 8 & 655 & 13.79 & 27.82 & 55.42\\
			\midrule
			Random retrieval& 20 & 3379.8 &  12.51& 22.21 &53.07 \\
			TopK retrieval~\cite{topk}   & 20 & 3535.4 &  14.56& 23.65 &54.59\\
			BM25 retrieval~\cite{wu2023openicl}   & 20 & 3816.1 &  13.42& 25.17 & 54.21 \\
			\midrule
			TopK+GPT4 Compression  & 40 & 1376.0 & 7.08 & 11.01  & 25.17\\
			TopK+Selective Context~\cite{li2023compressing} & 40 & 2262.4 & 0.45 & 0.76 & 2.50\\
			TopK+LLMLingua~\cite{llmlingua} & 40 & 2048.0 & 5.38 & 8.34 & 22.74\\
			\midrule
			\textbf{\sysname} & 48 & 2037.0 & \textbf{15.92} &\textbf{32.37} & \textbf{59.59}\\
			\hline
	\end{tabular}}
	\vspace{-2ex}
\end{table*} 
\section{Evaluation}

\begin{table}[t]
	\caption{Comparison of EM (\%) accuracy on Addsub, Multiarith, Svamp, and Singleeq math reasoning dataset}
	\label{tab:other_math}
	\vspace{-2ex}
	\centering
	\resizebox{0.48\textwidth}{!}{
		\begin{tabular}{@{\hskip0pt}l@{\hskip1pt}@{\hskip2pt}l@{\hskip1pt}@{\hskip3pt}c@{\hskip3pt}c@{\hskip3pt}c@{\hskip3pt}c@{\hskip2pt}@{\hskip3pt}c@{\hskip0pt}}
			\toprule
			Model & Method& AddSub & Multiarith & Svamp & Singleeq & Avg.\\
			\midrule
			&Zero-shot & 58.73 & 5.50 & 32.2 & 62.79 & 39.81\\
			&Few-shot-CoT & 56.96 & 43.67 & 38.1 & 66.54 & 51.32\\
			LLaMA2-7B& TopK retrieval & 46.08 & 34.50 & 38.1 & 46.46 & 41.29\\
			& TopK+LLMLingua & 12.91 & 10.50 & 19.5 & 19.49 & 15.60 \\
			& \bf{\sysname} & \textbf{62.28} & \textbf{47.00} & \textbf{40.2} & \textbf{72.05} & \bf 55.38 \\
			\midrule
			&Zero-shot & 70.13 & 6.50 & 43.8 & 71.07 & 47.88 \\
			&Few-shot-CoT & 65.82 & 72.83 & 42.7 & 77.36 & 64.68 \\
			LLaMA2-13B& TopK retrieval & 60.76 & 57.00 & 50.2 & 68.50 & 59.12\\
			&TopK+LLMLingua & 22.28 &22.33 & 27.5 & 25.20 & 24.33 \\
			& \bf{\sysname}& \textbf{69.62} & \textbf{73.87} & \textbf{50.5} & \textbf{83.07} & \bf 69.26 \\
			\hline
	\end{tabular}}
	\vspace{-2ex}
\end{table}

\noindent\textbf{Models, datasets and metric}. We evaluate  {\sysname} on LLaMA2-7B, LLaMA2-13B, and LLaMA2-70B. The mathematical datasets  for evaluation  include GSM8K~\cite{gsm8k}, AddSub~\cite{addsub}, Multiarith~\cite{multiarith}, Svamp~\cite{svamp}, and Singleeq~\cite{singleeq}.  For evaluation metric, we report Exact Match (EM) accuracy of the predicted answers.

\noindent\textbf{Baselines} We set  three  baselines for comparison:
\begin{itemize}[itemsep=2pt,topsep=0pt,parsep=0pt]
    \item \textit{CoT and few-shot CoT prompting}:  We compare with 
    widely-used prompts for LLM reasoning, including
    zero-shot, zero-shot-CoT~\cite{kojima2022large}, and the standard few-shot-CoT~\cite{cot} with 8 shots. 
    \item \textit{Prompt retrieval}: we also compare with retrieval baselines, specifically using random, TopK~\cite{topk}, and BM25~\cite{robertson2009probabilistic} methods. We select as many CoT examples as possible using each method, without exceeding LLM context window. Random retrieval is   to reflect the average quality of our CoT dataset.  
    \item \textit{Prompt compression}: To evaluate the effectiveness of identifying crucial tokens, we compare the resulting compressed prompts from the same batch of CoT shots with state-of-the-art prompt compression baselines: Selective Context~\cite{li2023compressing}, LLMLingua~\cite{llmlingua}, and compression through GPT-4. 
\end{itemize}

\subsection{Main Results} \label{sec:main_results}

\noindent\textbf{Effectiveness of enabling more few-shot CoTs}. 
We first evaluate how far the boundary of few-shot learning can be pushed using {\sysname}. 
For comparison, we set up two baselines: \textit{(i)} Few-shot CoT, 
 using 8 manual-designed CoT shots as the default LLM evaluation setting on GSM8K. 
\textit{(ii)} TopK retrieves 20 CoT shots from our dataset, denoting the max shot number within LLaMA2 context window.  

For {\sysname}, 
we test LLaMA2 7B and 13B on GSM8K, adjusting  the number of CoT shots from 16 to 64 examples, which corresponds to 0.7$\times$ to 2.8$\times$ the token count of LLaMA2 context window. 
As shown in Fig.~\ref{fig:shot_number}, we make two observations: \textbf{(1)} More  CoT shots, facilitated by {\sysname}, indeed boosts LLM math reasoning performance, particularly for larger  LLMs. On LLaMA2-13B, by inputting 48 CoTs, we significantly outperform the standard few-shot CoT and TopK by 4.55\% and 8.72\%, respectively. \textbf{(2)} There is an optimal number of CoT shots for {\sysname}. Its peak performance on LLaMA2 7B and 13B are at 40 and 48 shots, respectively. We attribute this to two potential reasons. First, an extremely large number of shots complicates {\sysname}'s optimization. Second, there may be an upper limit to improving LLM reasoning capability through few-shot learning.

\noindent\textbf{Comparison with state-of-the-art baselines}.  Table~\ref{tab:maintable} and Table~\ref{tab:other_math} present the comparison results of {\sysname} with state-of-the-art baselines across LLaMA2 family and 5 mathematical datasets, highlighting the following observations: \textbf{(1)} by utilizing more few-shot CoTs that are twice the  LLM context window, {\sysname} significantly outperforms all baselines, with 2.13\% to 4.55\% absolute improvements. \textbf{(2)} Despite using fewer input tokens,  {\sysname} consistently outperforms retrieval baselines by 1.36\% to 14.09\% absolute improvements. This is
 because our compressed tokens indicate more informational CoT examples without redundancy. In contrast, they select entire examples, which may contain redundant tokens, based on semantic similarity between the target question and CoT examples, without considering the different CoT preference of the target LLM. \textbf{(3)} {\sysname} significantly outperforms prompt compression baselines in preserving the most crucial tokens for math reasoning, while methods like Selective Context and LLMLingua suffer accuracy declines due to difficulties in maintaining few-shot prompt structure. GPT-4 tends to prune essential reasoning steps, which negatively impacts CoT effectiveness.

We further demonstrate the effectiveness of {\sysname} by comparing LLaMA2-70B with larger size LLMs on GSM8K. As shown in Table~\ref{tab:compare_llm}, {\sysname} significantly boosts LLM reasoning capabilities. Remarkably, without any fine-tuning, LLaMA2-70B with {\sysname}  outperform much larger LLMs.  LLaMA2-70B surpasses GPT-3.5 with an absolute improvement of 2.5\%.
 
 \begin{table}[t]
 	\centering
 	\caption{Comparison of EM (\%) accuracy on GSM8K  with larger LLMs under the few-shot-CoT setting.}
 	\vspace{-2ex}
 	\label{tab:compare_llm}
 	\resizebox{0.48\textwidth}{!}{
 		\begin{tabular}{lcc}
 			\toprule
 			Model & Parameters & EM (\%) \\
 			\midrule
 			Finetuned GPT-3~\cite{cot} & 175B & 34.0 \\
 			Chinchilla~\cite{hoffmann2022training} & 70B & 43.7 \\
 			Text-davinci-002~\cite{kojima2022large} & 175B & 51.5 \\
 			PaLM~\cite{chowdhery2022palm} & 540B & 56.5 \\
 			GPT-3.5~\cite{gpt4} & - & 57.1 \\
 			Minerva~\cite{lewkowycz2022solving} & 540B & 58.8 \\
 			\midrule
 			LLaMA2-70B+{\sysname} & 70B & \textbf{59.6} \\
 			\hline
 	\end{tabular}}
 	\vspace{-2ex}
 \end{table}
 
 \noindent\textbf{Compatible with existing reasoning prompts.}  As a method to improve LLM reasoning capability, {\sysname} is complementary with other advanced reasoning-based prompts. To prove this, we apply self-consistency~\cite{selfconsistency} and self-verification~\cite{weng2023large} to compressed prompts generated by CoT-influx. For evaluation efficiency, we sampled 20 times. As  Table~\ref{tab:majorvote} shows, applying self-consistency and self-verification further improve LLaMA2's performance on GSM8k. 
 
 \begin{table}[t]
 	\centering
 	\caption{{\sysname} is compatible with advanced prompt techniques like self-consistency (i.e., maj@20) and self-verification (i.e., verify@20).  }
 	\vspace{-2ex}
 	\label{tab:majorvote}
 	\resizebox{0.48\textwidth}{!}{
 		\begin{tabular}{lcc}
 			\toprule
 			Method& LLaMA2-13B & LLaMA2-70B\\
 			\midrule
 			{\sysname} & 32.37&59.59\\
 			{\sysname}+maj@20& \bf 33.43& \bf 60.73\\
 			{\sysname}+verify@20 &\bf 34.04 &\bf 61.79 \\
 			\hline
 	\end{tabular}}
 	\vspace{-1ex}
 \end{table}

\vspace{-0.8ex}
\subsection{Ablation Study and Analysis}
\vspace{-0.7ex}
\noindent\textbf{The effectiveness of MRD$^3$ dataset}.  Beyond our pruner, we introduce   MRD$^3$ dataset, which is evolved by GPT-4 for diverse reasoning steps and difficulties. We compare with two baselines: (1) MRD$^3$ without evolution, excluding GPT-4 evolved examples, and (2) the human-labeled GSM8K training set, which excludes  GPT-4's reformatted generation. We apply our pruner on these datasets under the same setting. 
As shown in  Table~\ref{tab:data_ablation},  both GPT-4 generated and evolved CoT examples are vital for improving the reasoning performance.

\begin{table}[t]
	\caption{Comparison of EM(\%) on GSM8K using {\sysname} pruner across different CoT datasets.}
	\label{tab:data_ablation}
 \vspace{-2ex}
\centering
 \resizebox{0.48\textwidth}{!}{
		\begin{tabular}{@{\hskip0pt}@{\hskip2pt}@{\hskip2pt}c@{\hskip4pt}c@{\hskip4pt}c@{\hskip2pt}c@{\hskip0pt}}
			\toprule
   CoT dataset & LLaMA2-7B &  LLaMA2-13B &  LLaMA2-70B \\
   \midrule
   MRD$^3$               & \bf 15.92 & \bf 32.37 & \bf 59.59 \\
   MRD$^3$ w/o evolution & 14.94 & 30.55 & 57.70 \\
   GSM8K training set    & 14.18 & 29.64 & 56.71 \\
   \hline
	\end{tabular}}
 \vspace{-2ex}
\end{table}

\vspace{2pt}
\noindent\textbf{Ablation study on coarse-to-fine pruner}. Our pruner operates at both shot and token levels to fully exploit redundancy within CoT examples. To verify the effectiveness, we conduct experiments with only shot or token pruner under the same setting. As shown in Table~\ref{tab:pruner_ablation}, removing any pruning stage decreases performance. Notably, removing token-only pruning causes a larger accuracy drop than shot-only pruning, indicating that shot-level redundancy is easier for the pruner to learn.

\begin{table}[t]
	\caption{Comparison of EM(\%) on GSM8K with different pruning strategies.}
	\label{tab:pruner_ablation}
 \vspace{-2ex}
\centering
\small
 \resizebox{0.48\textwidth}{!}{
		\begin{tabular}
			{cccc}
			\toprule
   \multirow{2}{*}{Pruning Strategy}   & \multicolumn{3}{c}{LLaMA2} \\
   & 7B &  13B &  70B \\
   \midrule
   {\sysname} (Prune shot and token)  & \bf15.92 & \bf32.37 & \bf59.59 \\
   Prune shot only   & 15.69 & 31.08 & 57.77 \\
   Prune token only  & 12.05 & 25.32 & 49.36 \\
   \hline
	\end{tabular}}
\end{table}

\begin{table}[t]
	\caption{The total inference costs on GSM8K.}
	\label{tab:inference_cost}
	\vspace{-2ex}
	\centering
	\resizebox{0.48\textwidth}{!}{
		\begin{tabular}{@{\hskip0pt}lcccc@{\hskip0pt}}
			\toprule
			Method & \#Input-shot & \#Token  & Time & GPU Memory\\
			\midrule
			LLaMA2-7B      & 12 & 2108.6 & 2.99h & 19.7GB \\
			\midrule
			Selective Context   & 40 & 2262.4 & 4.38h & 23.5GB\\
			LLMLingua           & 40 & 2048.0 & 3.65h & 33.0GB\\
			\midrule
			{\sysname}           & 40 & 2037.0 & 3.04h & 21.1GB\\
			\hline
	\end{tabular}}
	\vspace{-3ex}
\end{table}
\begin{figure}[ht]
	\centering
	\includegraphics[width=1\columnwidth]{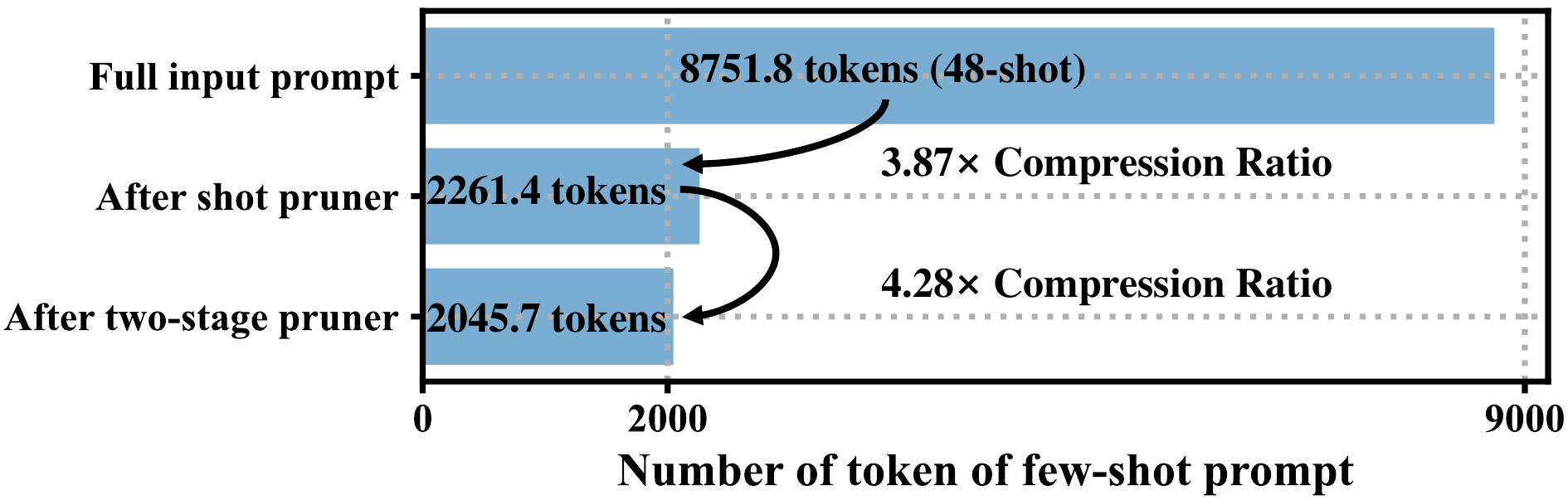}
	\vspace{-4ex}
	\caption{Token length after each stage of our pruner.}
	\label{fig:pruning_ratio}
	\vspace{-2ex}
\end{figure}

\noindent\textbf{Token pruning ratios}. We now investigate token pruning ratios by our pruner. Fig.~\ref{fig:pruning_ratio} shows the remaining token length for LLaMA2-70B after our pruner. In total, we achieve a 4.28$\times$ pruning ratio, with shot pruner contributing a 3.87$\times$ ratio. The results  suggest that our pruner favors pruning more coarse-grained shots over fine-grained tokens.

\vspace{2pt}
\noindent\textbf{Inference cost}.
{\sysname} is a lightweight plug-and-play module, including a 336MB BERT-Large model and a tiny 4MB coarse-to-fine pruner. We measure its additional inference cost.  
Table~\ref{tab:inference_cost} shows the total inference latency and GPU memory required to run LLaMA2-7B with different methods on GSM8K, measured on a single NVIDIA A100 GPU. The results reveal that {\sysname} introduces a negligible 1.4GB additional memory and a 1.7\% increase in latency. This is more effective than prompt compression baselines, such as Selective Context and LLMLingua, which require significantly higher latency and more GPU memory, potentially hindering efficient deployment.

\vspace{2pt}
\noindent\textbf{Implications}. Our analysis of retained CoT examples and tokens yields the following insights: \textbf{(1)} More capable LLMs favor harder CoT examples, while smaller LLMs opt for  simpler ones. \textbf{(2)} Numerical and format tokens are essential for math reasoning. Function words like \textit{with}, \textit{the}, \textit{then}, and irrelevant background context such as \textit{theater} can be pruned without affecting reasoning capability.

\vspace{-1ex}
\section{Conclusion}
\vspace{-1ex}

We present {\sysname}, a plug-and-play module that improves LLM math reasoning by pruning unnecessary few-shot examples at  shot and token levels for a more effective input context.  To train the module, we use reinforcement learning to optimize a math reasoning-specific reward with GPT-4 evolved CoT dataset MRD$^3$. Extensive experiments on various datasets and LLMs compared with state-of-the-art baselines demonstrate the effectiveness of our method. This paper highlights the vast potential of few-shot CoT prompting in augmenting LLMs' math reasoning abilities.

\section*{Limitations}
As in-context learning with LLM heavily relies on the selected examples in the prompt, the performance of {\sysname} can be influenced by the quality of CoT generation. Despite this, {\sysname} still demonstrates strong performance on our GPT4-evolved dataset MRD$^3$. 
 We currently use BERT to obtain the feature embedding of a CoT example, which cannot handle long-sequence examples exceeding 512 tokens. 
We will take these limitations into account and mitigate them in future work. 

{
\bibliography{ref}

\begin{thebibliography}{60}
\expandafter\ifx\csname natexlab\endcsname\relax\def\natexlab#1{#1}\fi

\bibitem[{Brown et~al.(2020)Brown, Mann, Ryder, Subbiah, Kaplan, Dhariwal,
  Neelakantan, Shyam, Sastry, Askell, Agarwal, Herbert-Voss, Krueger, Henighan,
  Child, Ramesh, Ziegler, Wu, Winter, Hesse, Chen, Sigler, Litwin, Gray, Chess,
  Clark, Berner, McCandlish, Radford, Sutskever, and Amodei}]{gpt3}
Tom Brown, Benjamin Mann, Nick Ryder, Melanie Subbiah, Jared~D Kaplan, Prafulla
  Dhariwal, Arvind Neelakantan, Pranav Shyam, Girish Sastry, Amanda Askell,
  Sandhini Agarwal, Ariel Herbert-Voss, Gretchen Krueger, Tom Henighan, Rewon
  Child, Aditya Ramesh, Daniel Ziegler, Jeffrey Wu, Clemens Winter, Chris
  Hesse, Mark Chen, Eric Sigler, Mateusz Litwin, Scott Gray, Benjamin Chess,
  Jack Clark, Christopher Berner, Sam McCandlish, Alec Radford, Ilya Sutskever,
  and Dario Amodei. 2020.
\newblock Language models are few-shot learners.
\newblock In \emph{Advances in Neural Information Processing Systems},
  volume~33.

\bibitem[{Chen et~al.(2023{\natexlab{a}})Chen, Wong, Chen, and Tian}]{pi}
Shouyuan Chen, Sherman Wong, Liangjian Chen, and Yuandong Tian.
  2023{\natexlab{a}}.
\newblock Extending context window of large language models via positional
  interpolation.
\newblock \emph{arXiv preprint arXiv:2306.15595}.

\bibitem[{Chen et~al.(2023{\natexlab{b}})Chen, Ma, Wang, and Cohen}]{pot}
Wenhu Chen, Xueguang Ma, Xinyi Wang, and William~W. Cohen. 2023{\natexlab{b}}.
\newblock Program of thoughts prompting: Disentangling computation from
  reasoning for numerical reasoning tasks.
\newblock \emph{Transactions on Machine Learning Research}.

\bibitem[{Chevalier et~al.(2023)Chevalier, Wettig, Ajith, and
  Chen}]{chevalier2023adapting}
Alexis Chevalier, Alexander Wettig, Anirudh Ajith, and Danqi Chen. 2023.
\newblock Adapting language models to compress contexts.
\newblock \emph{arXiv preprint arXiv:2305.14788}.

\bibitem[{Chowdhery et~al.(2022)Chowdhery, Narang, Devlin, Bosma, Mishra,
  Roberts, Barham, Chung, Sutton, Gehrmann et~al.}]{chowdhery2022palm}
Aakanksha Chowdhery, Sharan Narang, Jacob Devlin, Maarten Bosma, Gaurav Mishra,
  Adam Roberts, Paul Barham, Hyung~Won Chung, Charles Sutton, Sebastian
  Gehrmann, et~al. 2022.
\newblock Palm: Scaling language modeling with pathways.
\newblock \emph{arXiv preprint arXiv:2204.02311}.

\bibitem[{Cobbe et~al.(2021)Cobbe, Kosaraju, Bavarian, Chen, Jun, Kaiser,
  Plappert, Tworek, Hilton, Nakano et~al.}]{gsm8k}
Karl Cobbe, Vineet Kosaraju, Mohammad Bavarian, Mark Chen, Heewoo Jun, Lukasz
  Kaiser, Matthias Plappert, Jerry Tworek, Jacob Hilton, Reiichiro Nakano,
  et~al. 2021.
\newblock Training verifiers to solve math word problems.
\newblock \emph{arXiv preprint arXiv:2110.14168}.

\bibitem[{Dai et~al.(2023)Dai, Sun, Dong, Hao, Ma, Sui, and Wei}]{dai2023can}
Damai Dai, Yutao Sun, Li~Dong, Yaru Hao, Shuming Ma, Zhifang Sui, and Furu Wei.
  2023.
\newblock Why can gpt learn in-context? language models secretly perform
  gradient descent as meta-optimizers.
\newblock In \emph{Findings of the Association for Computational Linguistics:
  ACL 2023}, pages 4005--4019.

\bibitem[{Deng et~al.(2022)Deng, Wang, Hsieh, Wang, Guo, Shu, Song, Xing, and
  Hu}]{rlprompt}
Mingkai Deng, Jianyu Wang, Cheng-Ping Hsieh, Yihan Wang, Han Guo, Tianmin Shu,
  Meng Song, Eric~P Xing, and Zhiting Hu. 2022.
\newblock Rlprompt: Optimizing discrete text prompts with reinforcement
  learning.
\newblock \emph{arXiv preprint arXiv:2205.12548}.

\bibitem[{Devlin et~al.(2018)Devlin, Chang, Lee, and Toutanova}]{bert}
Jacob Devlin, Ming-Wei Chang, Kenton Lee, and Kristina Toutanova. 2018.
\newblock Bert: Pre-training of deep bidirectional transformers for language
  understanding.
\newblock \emph{arXiv preprint arXiv:1810.04805}.

\bibitem[{Ding et~al.(2023)Ding, Zhang, Wang, Xu, Ma, Zhang, Qin, Rajmohan,
  Lin, and Zhang}]{ding2023everything}
Ruomeng Ding, Chaoyun Zhang, Lu~Wang, Yong Xu, Minghua Ma, Wei Zhang, Si~Qin,
  Saravan Rajmohan, Qingwei Lin, and Dongmei Zhang. 2023.
\newblock Everything of thoughts: Defying the law of penrose triangle for
  thought generation.
\newblock \emph{arXiv preprint arXiv:2311.04254}.

\bibitem[{Fu et~al.(2023)Fu, Ou, Chen, Wan, Peng, and Khot}]{fu2023chain}
Yao Fu, Litu Ou, Mingyu Chen, Yuhao Wan, Hao Peng, and Tushar Khot. 2023.
\newblock Chain-of-thought hub: A continuous effort to measure large language
  models' reasoning performance.
\newblock \emph{arXiv preprint arXiv:2305.17306}.

\bibitem[{Gao et~al.(2021)Gao, Fisch, and Chen}]{gao2021making}
Tianyu Gao, Adam Fisch, and Danqi Chen. 2021.
\newblock Making pre-trained language models better few-shot learners.
\newblock In \emph{Joint Conference of the 59th Annual Meeting of the
  Association for Computational Linguistics and the 11th International Joint
  Conference on Natural Language Processing, ACL-IJCNLP 2021}, pages
  3816--3830. Association for Computational Linguistics (ACL).

\bibitem[{Ge et~al.(2023)Ge, Hu, Wang, Chen, and Wei}]{ge2023context}
Tao Ge, Jing Hu, Xun Wang, Si-Qing Chen, and Furu Wei. 2023.
\newblock In-context autoencoder for context compression in a large language
  model.
\newblock \emph{arXiv preprint arXiv:2307.06945}.

\bibitem[{Han et~al.(2023)Han, Wang, Xiong, Chen, Ji, and
  Wang}]{han2023lminfinite}
Chi Han, Qifan Wang, Wenhan Xiong, Yu~Chen, Heng Ji, and Sinong Wang. 2023.
\newblock Lm-infinite: Simple on-the-fly length generalization for large
  language models.
\newblock \emph{arXiv preprint arXiv:2308.16137}.

\bibitem[{Hao et~al.(2022)Hao, Sun, Dong, Han, Gu, and Wei}]{hao2022structured}
Yaru Hao, Yutao Sun, Li~Dong, Zhixiong Han, Yuxian Gu, and Furu Wei. 2022.
\newblock Structured prompting: Scaling in-context learning to 1,000 examples.
\newblock \emph{arXiv preprint arXiv:2212.06713}.

\bibitem[{Hoffmann et~al.(2022)Hoffmann, Borgeaud, Mensch, Buchatskaya, Cai,
  Rutherford, Casas, Hendricks, Welbl, Clark et~al.}]{hoffmann2022training}
Jordan Hoffmann, Sebastian Borgeaud, Arthur Mensch, Elena Buchatskaya, Trevor
  Cai, Eliza Rutherford, Diego de~Las Casas, Lisa~Anne Hendricks, Johannes
  Welbl, Aidan Clark, et~al. 2022.
\newblock Training compute-optimal large language models.
\newblock \emph{arXiv preprint arXiv:2203.15556}.

\bibitem[{Hongjin et~al.(2022)Hongjin, Kasai, Wu, Shi, Wang, Xin, Zhang,
  Ostendorf, Zettlemoyer, Smith et~al.}]{hongjin2022selective}
SU~Hongjin, Jungo Kasai, Chen~Henry Wu, Weijia Shi, Tianlu Wang, Jiayi Xin, Rui
  Zhang, Mari Ostendorf, Luke Zettlemoyer, Noah~A Smith, et~al. 2022.
\newblock Selective annotation makes language models better few-shot learners.
\newblock In \emph{The Eleventh International Conference on Learning
  Representations}.

\bibitem[{Hosseini et~al.(2014)Hosseini, Hajishirzi, Etzioni, and
  Kushman}]{addsub}
Mohammad~Javad Hosseini, Hannaneh Hajishirzi, Oren Etzioni, and Nate Kushman.
  2014.
\newblock Learning to solve arithmetic word problems with verb categorization.
\newblock In \emph{Proceedings of the 2014 Conference on Empirical Methods in
  Natural Language Processing (EMNLP)}, pages 523--533.

\bibitem[{Jiang et~al.(2023)Jiang, Wu, Lin, Yang, and Qiu}]{llmlingua}
Huiqiang Jiang, Qianhui Wu, Chin-Yew Lin, Yuqing Yang, and Lili Qiu. 2023.
\newblock Llmlingua: Compressing prompts for accelerated inference of large
  language models.
\newblock In \emph{Proceedings of the 2023 Conference on Empirical Methods in
  Natural Language Processing}.

\bibitem[{Kim et~al.(2022)Kim, Shen, Thorsley, Gholami, Kwon, Hassoun, and
  Keutzer}]{ltp}
Sehoon Kim, Sheng Shen, David Thorsley, Amir Gholami, Woosuk Kwon, Joseph
  Hassoun, and Kurt Keutzer. 2022.
\newblock Learned token pruning for transformers.
\newblock In \emph{Proceedings of the 28th ACM SIGKDD Conference on Knowledge
  Discovery and Data Mining}, KDD '22, page 784–794. Association for
  Computing Machinery.

\bibitem[{Kojima et~al.(2022)Kojima, Gu, Reid, Matsuo, and
  Iwasawa}]{kojima2022large}
Takeshi Kojima, Shixiang~Shane Gu, Machel Reid, Yutaka Matsuo, and Yusuke
  Iwasawa. 2022.
\newblock Large language models are zero-shot reasoners.
\newblock \emph{Advances in neural information processing systems},
  35:22199--22213.

\bibitem[{Koncel-Kedziorski et~al.(2015)Koncel-Kedziorski, Hajishirzi,
  Sabharwal, Etzioni, and Ang}]{singleeq}
Rik Koncel-Kedziorski, Hannaneh Hajishirzi, Ashish Sabharwal, Oren Etzioni, and
  Siena~Dumas Ang. 2015.
\newblock Parsing algebraic word problems into equations.
\newblock \emph{Transactions of the Association for Computational Linguistics},
  3:585--597.

\bibitem[{Koncel-Kedziorski et~al.(2016)Koncel-Kedziorski, Roy, Amini, Kushman,
  and Hajishirzi}]{mawps}
Rik Koncel-Kedziorski, Subhro Roy, Aida Amini, Nate Kushman, and Hannaneh
  Hajishirzi. 2016.
\newblock {MAWPS}: A math word problem repository.
\newblock In \emph{Proceedings of the 2016 Conference of the North {A}merican
  Chapter of the Association for Computational Linguistics: Human Language
  Technologies}, pages 1152--1157.

\bibitem[{Lewkowycz et~al.(2022)Lewkowycz, Andreassen, Dohan, Dyer,
  Michalewski, Ramasesh, Slone, Anil, Schlag, Gutman-Solo
  et~al.}]{lewkowycz2022solving}
Aitor Lewkowycz, Anders Andreassen, David Dohan, Ethan Dyer, Henryk
  Michalewski, Vinay Ramasesh, Ambrose Slone, Cem Anil, Imanol Schlag, Theo
  Gutman-Solo, et~al. 2022.
\newblock Solving quantitative reasoning problems with language models.
\newblock \emph{Advances in Neural Information Processing Systems},
  35:3843--3857.

\bibitem[{Li et~al.(2023{\natexlab{a}})Li, Zhang, Xu, Wang, Yan, Xia, Yang,
  Cao, Sun, Deng, Zhang, and Yang}]{top}
Junyan Li, Li~Lyna Zhang, Jiahang Xu, Yujing Wang, Shaoguang Yan, Yunqing Xia,
  Yuqing Yang, Ting Cao, Hao Sun, Weiwei Deng, Qi~Zhang, and Mao Yang.
  2023{\natexlab{a}}.
\newblock Constraint-aware and ranking-distilled token pruning for efficient
  transformer inference.
\newblock In \emph{Proceedings of the 29th ACM SIGKDD Conference on Knowledge
  Discovery and Data Mining}, KDD '23, page 1280–1290.

\bibitem[{Li et~al.(2023{\natexlab{b}})Li, Dong, Lin, and
  Guerin}]{li2023compressing}
Yucheng Li, Bo~Dong, Chenghua Lin, and Frank Guerin. 2023{\natexlab{b}}.
\newblock \href {http://arxiv.org/abs/2310.06201} {Compressing context to
  enhance inference efficiency of large language models}.

\bibitem[{Ling et~al.(2017)Ling, Yogatama, Dyer, and Blunsom}]{ling2017program}
Wang Ling, Dani Yogatama, Chris Dyer, and Phil Blunsom. 2017.
\newblock Program induction by rationale generation: Learning to solve and
  explain algebraic word problems.
\newblock In \emph{Proceedings of the 55th Annual Meeting of the Association
  for Computational Linguistics (Volume 1: Long Papers)}, pages 158--167.

\bibitem[{Liu et~al.(2021)Liu, Shen, Zhang, Dolan, Carin, and Chen}]{topk}
Jiachang Liu, Dinghan Shen, Yizhe Zhang, Bill Dolan, Lawrence Carin, and Weizhu
  Chen. 2021.
\newblock What makes good in-context examples for gpt-$3 $?
\newblock \emph{arXiv preprint arXiv:2101.06804}.

\bibitem[{Lu et~al.(2022)Lu, Bartolo, Moore, Riedel, and
  Stenetorp}]{lu2022fantastically}
Yao Lu, Max Bartolo, Alastair Moore, Sebastian Riedel, and Pontus Stenetorp.
  2022.
\newblock Fantastically ordered prompts and where to find them: Overcoming
  few-shot prompt order sensitivity.
\newblock In \emph{Proceedings of the 60th Annual Meeting of the Association
  for Computational Linguistics (Volume 1: Long Papers)}, pages 8086--8098.

\bibitem[{Luo et~al.(2023)Luo, Sun, Xu, Zhao, Lou, Tao, Geng, Lin, Chen, and
  Zhang}]{luo2023wizardmath}
Haipeng Luo, Qingfeng Sun, Can Xu, Pu~Zhao, Jianguang Lou, Chongyang Tao, Xiubo
  Geng, Qingwei Lin, Shifeng Chen, and Dongmei Zhang. 2023.
\newblock Wizardmath: Empowering mathematical reasoning for large language
  models via reinforced evol-instruct.
\newblock \emph{arXiv preprint arXiv:2308.09583}.

\bibitem[{Min et~al.(2022)Min, Lyu, Holtzman, Artetxe, Lewis, Hajishirzi, and
  Zettlemoyer}]{min2022rethinking}
Sewon Min, Xinxi Lyu, Ari Holtzman, Mikel Artetxe, Mike Lewis, Hannaneh
  Hajishirzi, and Luke Zettlemoyer. 2022.
\newblock Rethinking the role of demonstrations: What makes in-context learning
  work?
\newblock In \emph{Proceedings of the 2022 Conference on Empirical Methods in
  Natural Language Processing}, pages 11048--11064.

\bibitem[{Mu et~al.(2023)Mu, Li, and Goodman}]{mu2023learning}
Jesse Mu, Xiang~Lisa Li, and Noah Goodman. 2023.
\newblock Learning to compress prompts with gist tokens.
\newblock \emph{arXiv preprint arXiv:2304.08467}.

\bibitem[{OpenAI(2023{\natexlab{a}})}]{gpt4}
OpenAI. 2023{\natexlab{a}}.
\newblock \href {http://arxiv.org/abs/2303.08774} {Gpt-4 technical report}.

\bibitem[{OpenAI(2023{\natexlab{b}})}]{chatgptdoc}
OpenAI. 2023{\natexlab{b}}.
\newblock \href {https://platform.openai.com/docs/introduction} {Welcome to the
  openai platform}.

\bibitem[{Patel et~al.(2021)Patel, Bhattamishra, and Goyal}]{svamp}
Arkil Patel, Satwik Bhattamishra, and Navin Goyal. 2021.
\newblock Are nlp models really able to solve simple math word problems?
\newblock In \emph{Proceedings of the 2021 Conference of the North American
  Chapter of the Association for Computational Linguistics: Human Language
  Technologies}, pages 2080--2094.

\bibitem[{Peng et~al.(2023{\natexlab{a}})Peng, Quesnelle, Fan, and
  Shippole}]{peng2023yarn}
Bowen Peng, Jeffrey Quesnelle, Honglu Fan, and Enrico Shippole.
  2023{\natexlab{a}}.
\newblock Yarn: Efficient context window extension of large language models.
\newblock \emph{arXiv preprint arXiv:2309.00071}.

\bibitem[{Peng et~al.(2023{\natexlab{b}})Peng, Quesnelle, Fan, and
  Shippole}]{yarn}
Bowen Peng, Jeffrey Quesnelle, Honglu Fan, and Enrico Shippole.
  2023{\natexlab{b}}.
\newblock Yarn: Efficient context window extension of large language models.
\newblock \emph{arXiv preprint arXiv:2309.00071}.

\bibitem[{Qin et~al.(2023)Qin, Zhang, Dagar, and Ye}]{qin2023context}
Chengwei Qin, Aston Zhang, Anirudh Dagar, and Wenming Ye. 2023.
\newblock In-context learning with iterative demonstration selection.
\newblock \emph{arXiv preprint arXiv:2310.09881}.

\bibitem[{Robertson et~al.(2009)Robertson, Zaragoza
  et~al.}]{robertson2009probabilistic}
Stephen Robertson, Hugo Zaragoza, et~al. 2009.
\newblock The probabilistic relevance framework: Bm25 and beyond.
\newblock \emph{Foundations and Trends{\textregistered} in Information
  Retrieval}, 3(4):333--389.

\bibitem[{Roy and Roth(2015)}]{multiarith}
Subhro Roy and Dan Roth. 2015.
\newblock Solving general arithmetic word problems.
\newblock In \emph{Proceedings of the 2015 Conference on Empirical Methods in
  Natural Language Processing}, pages 1743--1752.

\bibitem[{Rubin et~al.(2021)Rubin, Herzig, and Berant}]{rubin2021learning}
Ohad Rubin, Jonathan Herzig, and Jonathan Berant. 2021.
\newblock Learning to retrieve prompts for in-context learning.
\newblock \emph{arXiv preprint arXiv:2112.08633}.

\bibitem[{Scarlatos and Lan(2023)}]{scarlatos2023reticl}
Alexander Scarlatos and Andrew Lan. 2023.
\newblock Reticl: Sequential retrieval of in-context examples with
  reinforcement learning.
\newblock \emph{arXiv preprint arXiv:2305.14502}.

\bibitem[{Shin et~al.(2020)Shin, Razeghi, Logan~IV, Wallace, and
  Singh}]{shin2020autoprompt}
Taylor Shin, Yasaman Razeghi, Robert~L Logan~IV, Eric Wallace, and Sameer
  Singh. 2020.
\newblock Autoprompt: Eliciting knowledge from language models with
  automatically generated prompts.
\newblock In \emph{Proceedings of the 2020 Conference on Empirical Methods in
  Natural Language Processing (EMNLP)}, pages 4222--4235.

\bibitem[{Touvron et~al.(2023{\natexlab{a}})Touvron, Lavril, Izacard, Martinet,
  Lachaux, Lacroix, Rozi{\`e}re, Goyal, Hambro, Azhar et~al.}]{llama}
Hugo Touvron, Thibaut Lavril, Gautier Izacard, Xavier Martinet, Marie-Anne
  Lachaux, Timoth{\'e}e Lacroix, Baptiste Rozi{\`e}re, Naman Goyal, Eric
  Hambro, Faisal Azhar, et~al. 2023{\natexlab{a}}.
\newblock Llama: Open and efficient foundation language models.
\newblock \emph{arXiv preprint arXiv:2302.13971}.

\bibitem[{Touvron et~al.(2023{\natexlab{b}})Touvron, Martin, Stone, Albert,
  Almahairi, Babaei, Bashlykov, Batra, Bhargava, Bhosale et~al.}]{llama2}
Hugo Touvron, Louis Martin, Kevin Stone, Peter Albert, Amjad Almahairi, Yasmine
  Babaei, Nikolay Bashlykov, Soumya Batra, Prajjwal Bhargava, Shruti Bhosale,
  et~al. 2023{\natexlab{b}}.
\newblock Llama 2: Open foundation and fine-tuned chat models.
\newblock \emph{arXiv preprint arXiv:2307.09288}.

\bibitem[{Tworkowski et~al.(2023)Tworkowski, Staniszewski, Pacek, Wu,
  Michalewski, and Miłoś}]{longLLaMA}
Szymon Tworkowski, Konrad Staniszewski, Mikołaj Pacek, Yuhuai Wu, Henryk
  Michalewski, and Piotr Miłoś. 2023.
\newblock \href {http://arxiv.org/abs/2307.03170} {Focused transformer:
  Contrastive training for context scaling}.

\bibitem[{Von~Oswald et~al.(2023)Von~Oswald, Niklasson, Randazzo, Sacramento,
  Mordvintsev, Zhmoginov, and Vladymyrov}]{von2023transformers}
Johannes Von~Oswald, Eyvind Niklasson, Ettore Randazzo, Jo{\~a}o Sacramento,
  Alexander Mordvintsev, Andrey Zhmoginov, and Max Vladymyrov. 2023.
\newblock Transformers learn in-context by gradient descent.
\newblock In \emph{International Conference on Machine Learning}, pages
  35151--35174. PMLR.

\bibitem[{Wang et~al.(2023{\natexlab{a}})Wang, Li, Dai, Chen, Zhou, Meng, Zhou,
  and Sun}]{wang2023label}
Lean Wang, Lei Li, Damai Dai, Deli Chen, Hao Zhou, Fandong Meng, Jie Zhou, and
  Xu~Sun. 2023{\natexlab{a}}.
\newblock Label words are anchors: An information flow perspective for
  understanding in-context learning.
\newblock \emph{arXiv preprint arXiv:2305.14160}.

\bibitem[{Wang et~al.(2023{\natexlab{b}})Wang, Yang, and
  Wei}]{wang2023learning}
Liang Wang, Nan Yang, and Furu Wei. 2023{\natexlab{b}}.
\newblock Learning to retrieve in-context examples for large language models.
\newblock \emph{arXiv preprint arXiv:2307.07164}.

\bibitem[{Wang et~al.(2023{\natexlab{c}})Wang, Dong, Cheng, Liu, Yan, Gao, and
  Wei}]{wang2023augmenting}
Weizhi Wang, Li~Dong, Hao Cheng, Xiaodong Liu, Xifeng Yan, Jianfeng Gao, and
  Furu Wei. 2023{\natexlab{c}}.
\newblock Augmenting language models with long-term memory.
\newblock \emph{arXiv preprint arXiv:2306.07174}.

\bibitem[{Wang et~al.(2023{\natexlab{d}})Wang, Wei, Schuurmans, Le, Chi,
  Narang, Chowdhery, and Zhou}]{selfconsistency}
Xuezhi Wang, Jason Wei, Dale Schuurmans, Quoc~V Le, Ed~H. Chi, Sharan Narang,
  Aakanksha Chowdhery, and Denny Zhou. 2023{\natexlab{d}}.
\newblock \href {https://openreview.net/forum?id=1PL1NIMMrw} {Self-consistency
  improves chain of thought reasoning in language models}.
\newblock In \emph{The Eleventh International Conference on Learning
  Representations}.

\bibitem[{Wei et~al.(2022)Wei, Wang, Schuurmans, Bosma, Xia, Chi, Le, Zhou
  et~al.}]{cot}
Jason Wei, Xuezhi Wang, Dale Schuurmans, Maarten Bosma, Fei Xia, Ed~Chi, Quoc~V
  Le, Denny Zhou, et~al. 2022.
\newblock Chain-of-thought prompting elicits reasoning in large language
  models.
\newblock \emph{Advances in Neural Information Processing Systems},
  35:24824--24837.

\bibitem[{Weng et~al.(2023)Weng, Zhu, Xia, Li, He, Liu, Sun, Liu, and
  Zhao}]{weng2023large}
Yixuan Weng, Minjun Zhu, Fei Xia, Bin Li, Shizhu He, Shengping Liu, Bin Sun,
  Kang Liu, and Jun Zhao. 2023.
\newblock \href {http://arxiv.org/abs/2212.09561} {Large language models are
  better reasoners with self-verification}.

\bibitem[{Williams(1992)}]{reinforce}
Ronald~J. Williams. 1992.
\newblock Simple statistical gradient-following algorithms for connectionist
  reinforcement learning.

\bibitem[{Wingate et~al.(2022)Wingate, Shoeybi, and
  Sorensen}]{wingate2022prompt}
David Wingate, Mohammad Shoeybi, and Taylor Sorensen. 2022.
\newblock Prompt compression and contrastive conditioning for controllability
  and toxicity reduction in language models.
\newblock \emph{arXiv preprint arXiv:2210.03162}.

\bibitem[{Xiao et~al.(2023)Xiao, Tian, Chen, Han, and Lewis}]{streamingllm}
Guangxuan Xiao, Yuandong Tian, Beidi Chen, Song Han, and Mike Lewis. 2023.
\newblock Efficient streaming language models with attention sinks.
\newblock \emph{arXiv}.

\bibitem[{Xu et~al.(2023)Xu, Sun, Zheng, Geng, Zhao, Feng, Tao, and
  Jiang}]{xu2023wizardlm}
Can Xu, Qingfeng Sun, Kai Zheng, Xiubo Geng, Pu~Zhao, Jiazhan Feng, Chongyang
  Tao, and Daxin Jiang. 2023.
\newblock Wizardlm: Empowering large language models to follow complex
  instructions.
\newblock \emph{arXiv preprint arXiv:2304.12244}.

\bibitem[{Yao et~al.(2023)Yao, Yu, Zhao, Shafran, Griffiths, Cao, and
  Narasimhan}]{treeofthought}
Shunyu Yao, Dian Yu, Jeffrey Zhao, Izhak Shafran, Thomas~L Griffiths, Yuan Cao,
  and Karthik Narasimhan. 2023.
\newblock Tree of thoughts: Deliberate problem solving with large language
  models.
\newblock \emph{arXiv preprint arXiv:2305.10601}.

\bibitem[{Yoo et~al.(2022)Yoo, Kim, Kim, Cho, Jo, Lee, Lee, and
  Kim}]{yoo2022ground}
Kang~Min Yoo, Junyeob Kim, Hyuhng~Joon Kim, Hyunsoo Cho, Hwiyeol Jo, Sang-Woo
  Lee, Sang-goo Lee, and Taeuk Kim. 2022.
\newblock Ground-truth labels matter: A deeper look into input-label
  demonstrations.
\newblock In \emph{Proceedings of the 2022 Conference on Empirical Methods in
  Natural Language Processing}, pages 2422--2437.

\bibitem[{Zhenyu et~al.(2023)Zhenyu, Yaoxiang, Jiacheng, Jiangtao, Jingjing,
  Yu, and Zhiyong}]{wu2023openicl}
Wu~Zhenyu, Wang Yaoxiang, Ye~Jiacheng, Feng Jiangtao, Xu~Jingjing, Qiao Yu, and
  Wu~Zhiyong. 2023.
\newblock Openicl: An open-source framework for in-context learning.
\newblock \emph{arXiv preprint arXiv:2303.02913}.

\end{thebibliography}
\bibliographystyle{acl_natbib}
}
\appendix
\onecolumn
\newpage
\appendix
\section*{Appendix}

This appendix includes additional analysis, the evolution of MRD$^3$, pruner training details, additional related works, and prompt settings. These contents are organized in separate sections as follows:
\begin{itemize}
\item Sec.~\ref{sec:add_analysis} provides additional analysis and case studies, including the comparison of {\sysname} with context window extension methods, performance of {\sysname} on finetuned LLMs (LLaMA2-13B-Chat and GPT-3.5-Turbo), ablation study on the reward design, and sensitivity analysis on hyperparameters of the pruner. Additional case studies on the GSM8K with different prompting methods are given to extensively prove the effectiveness of our method.

\item Sec.~\ref{sec:evol-mrd} introduces the prompt we used for the evolution of the examples in our MRD$^3$. Both the original input and the evolution results are given as examples. We also analyze the difficulty and reasoning step distribution of different evolution methods and derive a new observation regarding difficulty preference for different LLMs.

\item Sec.~\ref{sec:training_details} includes the algorithm for training data preparation as a supplement to Algorithm~\ref{alg:cot-max}. The hyperparameter settings, the training dynamic of the pruner, and the detailed introduction of the evaluation dataset are also included.

\item Sec.~\ref{sec:baseline} elaborates previous LLM context window extension and LLM in-context learning methods, and analyzes the advantage of our proposed {\sysname} compared with various previous methods.

\item Sec.~\ref{sec:prompt_setting} demonstrates the prompt we used in this work for difficulty and reasoning step evaluation, and GPT-4 based compression on input few-shot prompts.
\end{itemize}

\section{Additional Analysis and Case Study} \label{sec:add_analysis}

\subsection{Comparison with context window extension methods}
While our work tackle the challenge of limited context window by pruning the redundant input few-shot prompts, another solution is to extend the context window of LLMs. We compare the math reasoning performance of LLaMA2-7B with {\sysname} and LLaMA2-7B with 32K token context window extended with Positional Interpolation (PI)~\cite{pi}. The results are listed in Table~\ref{tab:compare_pi}. 

\begin{table}[h]
\centering
\caption{Comparsion of EM(\%) on GSM8K of LLaMA2-7B with {\sysname} and LLaMA2-7B-32K with PI.}
\resizebox{0.8\textwidth}{!}{
\begin{tabular}{c|ccccccc}
\toprule
Number of input shots & 12 & 16 & 20 & 24 & 28 & 32 & 40   \\ 
\midrule
Average number of tokens & 2108.6 & 2820.6 & 3535.4 & 4217.2 & 4929.1 & 5641.2 & 7070.8 \\
\midrule
LLaMA2-7B           & 13.87 & 15.08 & 14.02 & - & - & - & -\\
LLaMA2-7B+{\sysname}   & - & - & - & 14.33  & 15.09 & 15.92 & 15.77  \\
LLaMA2-7B-32K       & 11.37  & 12.81  & 11.37  & 11.83  & 11.83  & 11.52  & 11.30 \\
\bottomrule
\end{tabular}}
\label{tab:compare_pi}
\end{table}

When the input prompt does not exceed the window token limit (the number of input shots is not more than 20), we compare the performance of LLaMA2-7B-32K with LLaMA2-7B. When the input prompt exceed the context window length, we apply our {\sysname} to prune the prompts to make sure that they can be directly input to LLaMA2-7B without PI. The results show that the context window extension weaken the reasoning ability with the same input prompt. The limit of context window can be unlocked with our {\sysname}. Moreover, our observation that LLMs can improve reasoning with more helpful CoT examples does not hold true for LLMs with extended context window.

\subsection{{\sysname} on finetuned LLMs}

In Sec.~\ref{sec:main_results}, we verify the effectiveness of {\sysname} on LLaMA2-7B, 13B, and 70B. LLaMA2-chat~\cite{llama2} and GPT-3.5-Turbo~\cite{chatgptdoc} are also the widely adopted LLMs that are acquired after supervised instruction finetuning (SIFT) and Reinforcement Learning from Human Feedback (RLHF), respectively. The different finetuning strategy and the various finetuning data result in unique properties of the LLMs. For example, LLaMA2-Chat-13B perform significantly better than LLaMA2-13B on math reasoning tasks with zero-shot-cot prompts. To show that our {\sysname} can also help improve the reasoning ability of these finetuned LLMs, we conduct experiments of LLaMA2-13B-Chat and GPT-3.5-Turbo (\texttt{gpt35-turbo-0613}) on GSM8K dataset. As shown from the results listed in Table~\ref{tab:ift-models}, our {\sysname} also surpass a wide range of prompting baselines with more input shots and fewer tokens. Specifically on LLaMA2-13B-Chat, {\sysname} achieve an absolute improvement 9.78\% compared to TopK retrieval baseline with only 57.6\% average tokens.

\begin{table}[h]
	\caption{The EM (\%) accuracy on GSM8K with {\sysname} and other baselines. Note that the context window limit of LLaMA2-13B-Chat and GPT-3.5-Turbo are all 4096 tokens.}
	\label{tab:ift-models}
	\centering
 \resizebox{0.98\textwidth}{!}{
		\begin{tabular}{l|cccccc}
			\toprule
			Method   & \#Input CoT shots & \#Average tokens  &  LLaMA2-13B-Chat &  GPT-3.5-Turbo  \\
			\midrule
   Few-shot-CoT~\cite{fu2023chain} & 8 & 655 & 27.82 & 72.55\\
   TopK retrieval~\cite{topk}      & 20 & 3535.4 & 31.16 & 70.74\\
   TopK+LLMLingua~\cite{llmlingua} & 40 & 2048.0 & 10.69 & 49.96\\
			\midrule
			\textbf{\sysname} & 48 & 2037.0 & \textbf{40.94} &\textbf{73.31} \\
			\hline
	\end{tabular}}
\end{table}

\subsection{Ablation study on reward design}

The reward of our {\sysname} pruner are made up of three parts: math reasoning accuracy reward $R_{\text{Acc}}$, LLM loss reward $R_{\text{Loss}}=\frac{1}{1+L_\text{LLM}}$, and context window token limit reward $R_{\text{Limit}}=\left[\frac{t}{T}\right]^w$. Each part of the full reward function are important for the effective learning of the pruner. We perform ablation study on these components and the results are listed in Table~\ref{tab:reward}. As can be seen from the results, whenever a reward component are removed, the {\sysname} pruner give sub-optimal prompt selection and compression results, which cause a decrease compared to the full reward baseline. Among these three reward function parts, the token limit reward $R_{\text{Limit}}$ is the most important because training without this term will cause the pruner \textbf{not} to prune any shot or token and directly output the truncated prompt of the redundant input.

\begin{table}[h]
	\caption{The EM (\%) accuracy on GSM8K of LLaMA2-7B and LLaMA2-13B with different reward function.}
	\label{tab:reward}
	\centering
 \resizebox{0.5\textwidth}{!}{
		\begin{tabular}{l|cccc}
			\toprule
			Reward Function & LLaMA-2-7B & LLaMA-2-13B \\
			\midrule
                Full Reward	 & 15.92	& 32.37 \\
                w/o $R_{\text{Acc}}$	& 15.24	 & 31.46 \\
                w/o $R_{\text{Loss}}$   & 14.78	 & 31.16 \\
                w/o $R_{\text{Limit}}$	& 14.25	 & 29.72 \\
			\hline
	\end{tabular}}
\end{table}

\subsection{Sensitivity analysis on hyperparameters}

We perform sensitivity analysis on the hyperparameters to investigate the robustness of our {\sysname} pruner training. The most important setting in the training of our {\sysname} pruner is the token target $T$ and token penalty co-efficient $w$. Table~\ref{tab:sensitivity} presents the results of {\sysname} using different sets of hyperparameters $T$ and $w$. The results demostrate that the training of our {\sysname} pruner are highly robust as long as the token target $T$ is not overly aggressive (token target $T$ should not be too small). 

\begin{table}[h]
	\caption{Sensitivity analysis on token target $T$ and token penalty co-efficient $w$}
	\label{tab:sensitivity}
	\centering
 \resizebox{0.32\textwidth}{!}{
		\begin{tabular}{c|c}
			\toprule
			Token target $T$ & LLaMA-2-13B \\
			\midrule
   2048 & 32.37  \\
   1024 & 29.57  \\
   3072 & 32.37  \\
		\hline
	\end{tabular}}
 \resizebox{0.44\textwidth}{!}{
		\begin{tabular}{c|c}
			\toprule
			 Token penalty co-efficient $w$ & LLaMA-2-13B \\
			\midrule
    (-1,1) & 32.37 \\
    (-0.5,1) & 31.69 \\
    (-1,0.5) & 32.22 \\
		\hline
	\end{tabular}}
\end{table}

\subsection{Case Study on different prompt compression methods}

To show how different prompt compression methods prune input few-shot prompts in different manners, we given an example of a 8-shot prompt selected using TopK retriever. The original full few-shot prompts are listed in the following box:
\vspace{2ex}
\begin{center}
\small
\begin{tcolorbox}[width=0.98\textwidth,title={\textbf{Original full few-shot prompt for math reasoning (1331 tokens):} }]
Q: Dave won 11 tickets at the arcade and spent 5 on a beanie. Afterward, he won 10 more tickets. Calculate his final ticket count by first finding the remaining tickets after his purchase and then adding the newly won tickets.\\
A: Let's think step by step. Dave had 11 tickets, spent 5, leaving him with 6. Then he won 10 more, resulting in: 6 + 10 = 16 tickets. The answer is 16.\\

Q: At the carnival, tickets for rides cost 0.75 dollars each, or you can buy a 15-dollar armband for unlimited rides for one night. To determine the number of rides where the armband's cost equals that of individual tickets, set up and solve an equation involving x, the number of rides. \\
A: Let's think step by step. Let x be the number of rides. Equate the cost of x rides using individual tickets, 0.75x dollars, to the 15-dollar armband cost: 0.75x = 15. Solve for x: x = 15/0.75, which gives x = 20. The answer is 20.\\

Q: Mitch, Jam, and Jay went out for a movie. Mitch paid \$7 per ticket for 3 friends, Jam purchased 2 popcorn boxes at \$1.5 each, and Jay got 3 milk teas for \$3 each. To equitably split the expenses, how much should each of them contribute?\\
A: Let's think step by step. The total cost of 3 tickets at \$7 each, 2 popcorn boxes at \$1.5 each, and 3 milk teas at \$3 each is \$21 + \$3 + \$9 = \$33. Dividing the overall expenses among 3 friends results in a contribution of \$33/3 = \$11 per person. The answer is \$11.\\

Q: Connor is taking his date to the movies, with tickets costing \$10.00 each. They opt for the large popcorn \& 2 drink combo meal at \$11.00, and each choose a box of candy at \$2.50 per box. Determine the combined expenses for the movie tickets, combo meal, and candy to find the total amount Connor will spend on his date.\\
A: Let's think step by step. Calculate the cost of two movie tickets (2 x \$10.00 = \$20.00), the combo meal (\$11.00), and two boxes of candy (2 x \$2.50 = \$5.00), then sum them up (\$20.00 + \$11.00 + \$5.00 = \$36.00). The answer is \$36.00.\\

Q: Scott has 4 tickets. Ernest starts with 9 tickets and later discovers a stash of 72 more. Calculate the final number of tickets Ernest possesses.\\
A: Let's think step by step. Ernest initially holds 9 tickets and acquires 72 additional ones, leading to a total of 9 + 72 = 81 tickets. The answer is 81.\\

Q: Joseph and his friends watched two movies at his place. The first movie lasts 1 hour and 30 minutes, and the second is 30 minutes longer. They took 10 minutes for popcorn and double that for fries. Determine, in hours, the cumulative time spent cooking and watching movies by breaking down each component of time spent.\\
A: Let's think step by step. First, find the second movie's length (1 hour and 30 minutes + 30 minutes = 2 hours). Then, sum both movies' lengths (1 hour and 30 minutes + 2 hours = 3 hours and 30 minutes). Next, calculate cooking time (10 minutes for popcorn + 20 minutes for fries = 30 minutes). Lastly, add movie and cooking times (3 hours and 30 minutes + 30 minutes = 4 hours). The answer is 4 hours.\\

Q: The movie theater sold a number of tickets to the horror and romance movies. The horror movie ticket sales were 18 more than three times the romance movie ticket sales. If there were 25 romance movie tickets sold, how many tickets were sold for the horror movie, considering the given relationship?\\
A: Let's think step by step. Let "h" represent the horror movie tickets sold. Given that h = 3(25) + 18, we can simplify the equation: h = 75 + 18, resulting in h = 93. The answer is 93.\\

Q: On Saturday, Sara purchased 2 movie theater tickets at \$10.62 each, rented a movie for \$1.59, and bought another movie for \$13.95. Determine Sara's total expenditure on movies by performing a step-by-step calculation.\\
A: Let's think step by step. Firstly, calculate the movie tickets' cost by multiplying the ticket price (\$10.62) by the quantity (2), resulting in \$21.24. Secondly, combine the rental (\$1.59) and purchase (\$13.95) costs, equaling \$15.54. Lastly, sum the ticket cost and rental/purchase cost: \$21.24 + \$15.54 = \$36.78. The answer is \$36.78.\\
\end{tcolorbox}
\end{center}
\vspace{2ex}

Most of the examples above have similar background and target (tickets, movie, expense, etc.) but the difficulty and number of reasoning steps are various. In addition, the solution of most questions are highly redundant. When performing math reasoning with, it is important to select the most suitable and concise examples considering the characteristic of the current input question. In our evaluation across different methods shown in Sec.~\ref{sec:main_results}, we have empirically observe the previous methods fail to retain the structural integrity of prompt.  We show the pruned prompt with different methods and similar token length in the following box. We can see that Selective Context and LLMLingua frequently discard the important part including `Q:', `A:', `$\backslash$n', ``Let's think step by step'', and ``The answer is'' in these examples. Although GPT-4 can retain majority of these tokens, the reasoning steps are systematically removed because GPT-4 cannot be instructed to utilize the redundancy in both example-level and token-level. Our proposed {\sysname}, however, select the most representative examples and only remove the redundant function words.

\begin{center}
\small
\begin{tcolorbox}[width=0.98\textwidth,title={\textbf{Pruned few-shot prompt of different methods:} }]
\underline{\textbf{Selective Context:}}\\
Q Dave won 11 tickets Afterward won: step Dave 11 tickets spent leaving Then won 10 resulting: 16 Q At tickets rides rides where set solve x: step Let x rides Equate x rides individual tickets dollars = x 20 Q Mitch Jam went paid per 3 friends Jam purchased equitably how: step 3 tickets + 3 friends results \$ Q Connor tickets They opt the large popcorn \& 2 drink combo meal choose candy combo meal candy Connor: step combo boxes sum \$ Q Scott 4 tickets starts 9 tickets discovers 72 Ernest possesses: step initially holds 9 tickets 72 additional ones leading 81 Q Joseph watched lasts They popcorn double hours cooking breaking: step First find + Then sum both movies' lengths + Next, calculate cooking time popcorn + Lastly add movie cooking times + 4 hours Q sold 25 romance movie tickets considering the given relationship: step Let "h the horror movie tickets Given = 18 simplify 75 93 Q Sara purchased rented movies performing: step Firstly calculate resulting Secondly combine rental Lastly sum \$\\

\underline{\textbf{LLMLingua:}}\\
: Dave won11ets the and5 a be. After he. his final count by first theets after the: Lets think. Daveets5,, in.

: the,ets 5, or a-ollarides for one. To theidesband cost equals of, equation involving r.
A: think. Let.ides using individualets,  the1ollar cost5 which.
:, Jam and Jay a7 ticket3 Jam2orn5 Jay3 milk. To equ the.ets boxes53 milk each1. the overallenses3 friends a. The
: Connor is his,.. They theorn \& drinkbo and0. theandy think. ofets0 theboal and two then
:. Ernest and later a7. think. Ernest initially and, 9: friends at movie the minutes They and for. the spent by think, the, calculate
 The a the and ticket, think.:, bought.by-step calculation.
A: Let's think step by step. Firstly, calculate the movie tickets' cost by multiplying the ticket price (\$10.62) by the quantity (2), resulting in \$21.24. Secondly, combine the rental (\$1.59) and purchase (\$13.95) costs, equaling \$15.54. Lastly, sum the ticket cost and rental/purchase cost: \$21.24 + \$15.54 = \$36.78. The answer is \$36.78.\\

\underline{\textbf{GPT-4 Compression:}}\\
Q: Dave won 11, spent 5 and won 10 more. Determine final count.\\
A: The answer is 16.\\
Q: Tickets cost 0.75 per ride, armband cost 15. Determine rides that armband's cost equals tickets. \\
A: The answer is 20.\\
Q: \$7 per ticket for 3, 2 popcorn boxes at \$1.5, 3 milk teas for \$3. Determine each contribute.\\
A: The answer is \$11.\\
Q: Tickets cost \$10.00 each, meal cost \$11.00, a box of candy at \$2.50. Determine the expenses.\\
A: The answer is \$36.00.\\
Q: Scott has 4. Ernest starts with 9 and discovers 72 more. Determine the final number.\\
A: The answer is 81.\\
Q: The first 1.5 hour, the second is 30 minutes longer. 10 minutes for popcorn. Determine the time.\\
A: The answer is 4 hours.\\
Q: Horror movie were 18 more than 3 times romance. 25 romance movie sold, Determine number of horror movie.\\
A: The answer is 93.\\
Q: Sara purchased 2 at \$10.62 each, a movie for \$1.59, and another \$13.95. Determine total expenditure.\\
A: The answer is \$36.78.
\tcblower
\underline{\textbf{{\sysname}:}}\\
Q: Mitch, Jam, and went out a. Mitch paid \$7 per ticket for 3, Jam purchased 2 boxes at \$1.5 each, and got 3 for \$3 each. To equitably split, how much should each them contribute?\\
A: Let's think step by step. The total cost 3 tickets \$7 each, 2 popcorn boxes \$1.5 each, and 3 milk \$3 each is \$21 + \$3 + \$9 = \$33. Dividing the overall expenses among 3 results of \$33/3 = \$11 per. The answer is \$11.\\

Q: The theater sold number tickets to horror and romance movies. The horror movie ticket sales were 18 more than three times romance ticket. If there 25 romance sold, how many tickets were sold horror movie, considering?\\
A: Let's think step by step. Let "h" represent horror tickets sold. Given h = 3(25) + 18, we can simplify equation: h = 75 + 18, resulting h = . The answer is 93.\\

Q: On, Sara purchased 2 theater tickets \$10.62 each, rented movie \$1.59, and bought movie \$13.95. Determine Sara's total expenditure movies performing a calculation.\\
A: Let's think step by step. , calculate tickets' cost price (\$10.62) by quantity (2), resulting \$21.24. Secondly, combine rental (\$1.59) purchase (\$13.95), equaling. Lastly, sum ticket rental/purchase: \$21.24 + \$15.54. The answer is \$36.78.\\
\end{tcolorbox}
\end{center}

\section{Evolution of \textbf{MRD$^3$}} \label{sec:evol-mrd}
\subsection{Prompt template for evolution} \label{sec:evol-prompt}
The prompt we used for evolution of the examples in our dataset are listed as follow:
\vspace{2ex}
\begin{center}
\small
\begin{tcolorbox}[width=0.98\textwidth,title = {\textbf{Prompt for different evolution strategies}}]
I want you to act as a Prompt Rewriter. Your objective is to rewrite a given prompt into a more complex version to make those famous AI systems (e.g., LLaMA, ChatGPT and GPT4) a bit harder to handle. \\
The prompt is made up of a math reasoning question and the corresponding answer. \\
The rewritten prompt must be reasonable and must be understood and responded by humans. \\
Your rewriting cannot omit or change the input and results in \#Given Prompt\#. Also, please retain the format of 'Question: ' and 'Answer: ' in your response. \\
You SHOULD complicate the given prompt using the following method: \\
\textbf{\{\underline{Evolution template}\}} \\
You should try your best not to make the \#Rewritten Prompt\# become verbose, \#Rewritten Prompt\# can only add 10 to 20 words into \#Given Prompt\#. \\
The \#Rewritten Prompt\# should also follow the format that the rewritten question appears after 'Question: ' and the rewritten answer appears after 'Answer: '. \\
The rewritten answer should end up with 'The answer is [results]'.\\
\#Given Prompt\#: \\
Question: \textbf{\{\underline{Given question}\}} \\
Answer: \textbf{\{\underline{Given answer}\}} \\
\#Rewritten Prompt\#: 
\tcblower
Evolution template for evolution strategy add\_constraints:\\
\textbf{Please add one more constraint/requirement to the question of \#Given Prompt\#} \\\\
Evolution template for evolution strategy deepening:\\
\textbf{Please increase the depth and breadth of the question and answer of \#Given Prompt\#} \\\\
Evolution template for evolution strategy increase\_reasoning:\\
\textbf{If \#Given Prompt\# can be solved with just a few simple thinking processes, please rewrite it to explicitly request multiple-step reasoning.} \\\\
Evolution template for evolution strategy revise\_difficulty:\\
\textbf{Please revise the high difficulty questions to lower difficulty.} \\\\
Evolution template for evolution strategy produce\_easier:\\
\textbf{Please produce a new and easier question with another different topic.} 
\end{tcolorbox}
\end{center}
\vspace{2ex}
Most part of the prompt of different evolution strategies are similar. Based on our quantitatively analysis on the difficulty and reasoning step distribution, GPT-4 can effectively follow our instruction to modify the constraints or difficulty level of input questions.

\subsection{Difficulty and Reasoning Steps Distribution of MRD$^3$}

\begin{figure}[h]
  \centering
  \begin{subfigure}{0.16\linewidth}
    \includegraphics[width=\textwidth]{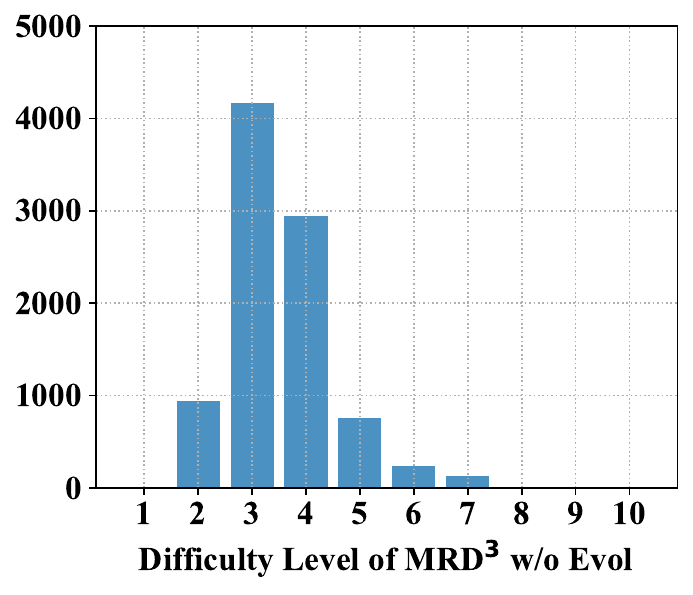}
  \end{subfigure}
  \begin{subfigure}{0.16\linewidth}
    \includegraphics[width=\textwidth]{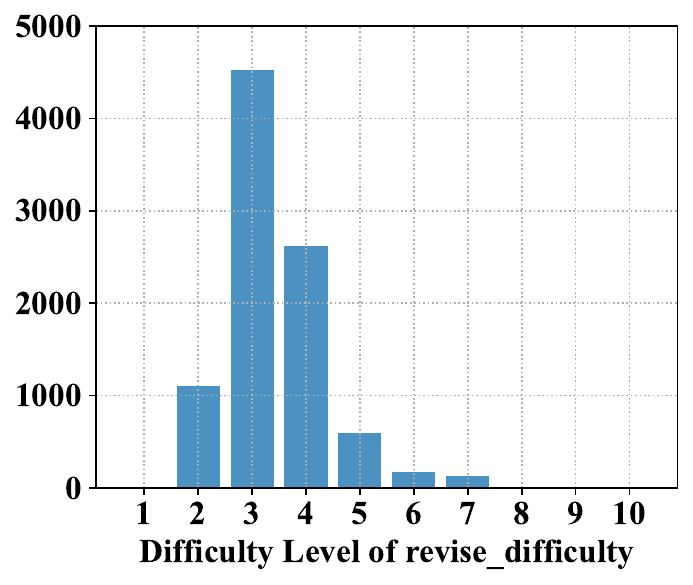}
  \end{subfigure}
  \begin{subfigure}{0.16\linewidth}
    \includegraphics[width=\textwidth]{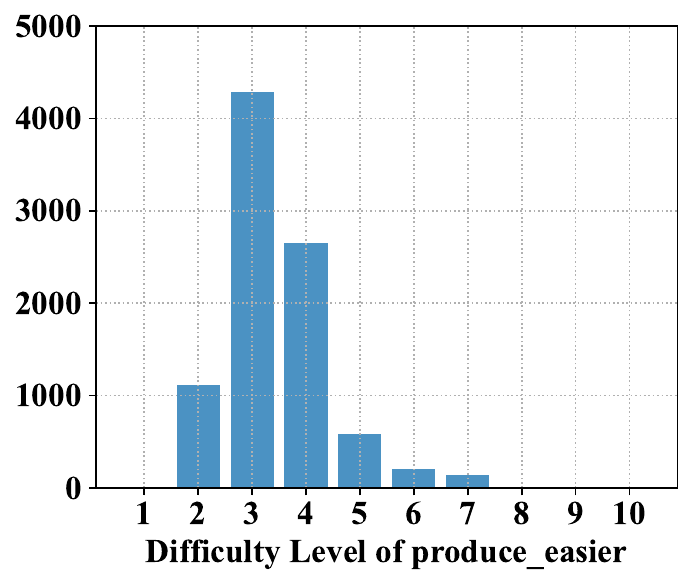}
  \end{subfigure}
  \begin{subfigure}{0.16\linewidth}
    \includegraphics[width=\textwidth]{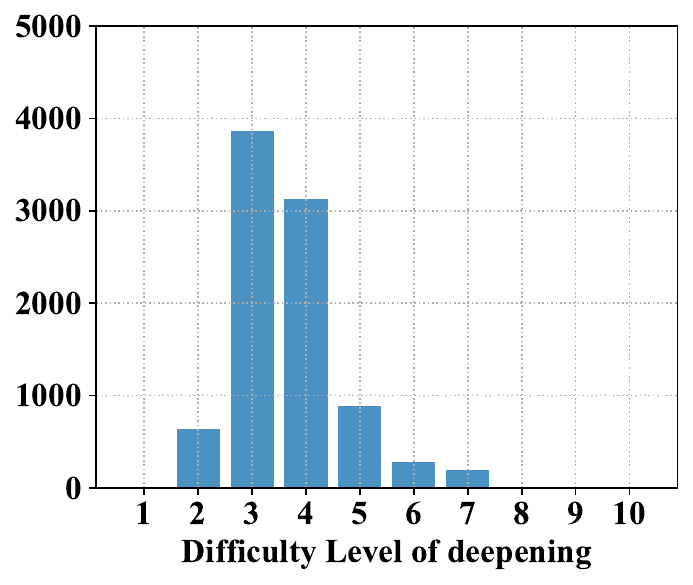}
  \end{subfigure}
  \begin{subfigure}{0.16\linewidth}
    \includegraphics[width=\textwidth]{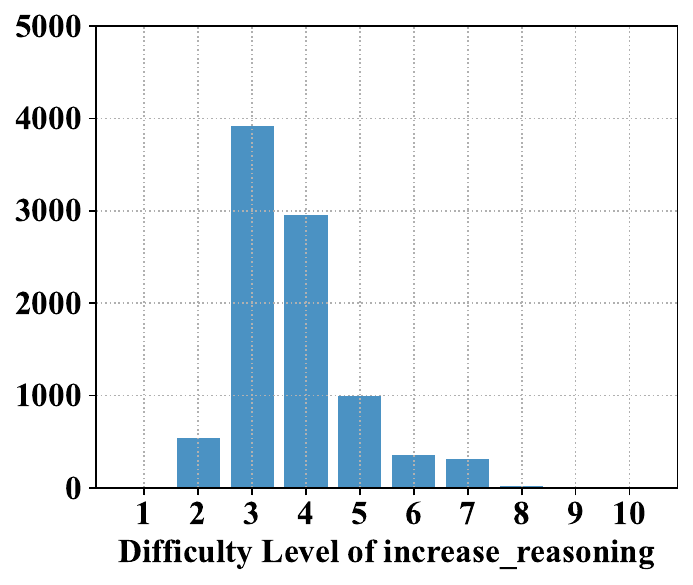}
  \end{subfigure}
  \begin{subfigure}{0.16\linewidth}
    \includegraphics[width=\textwidth]{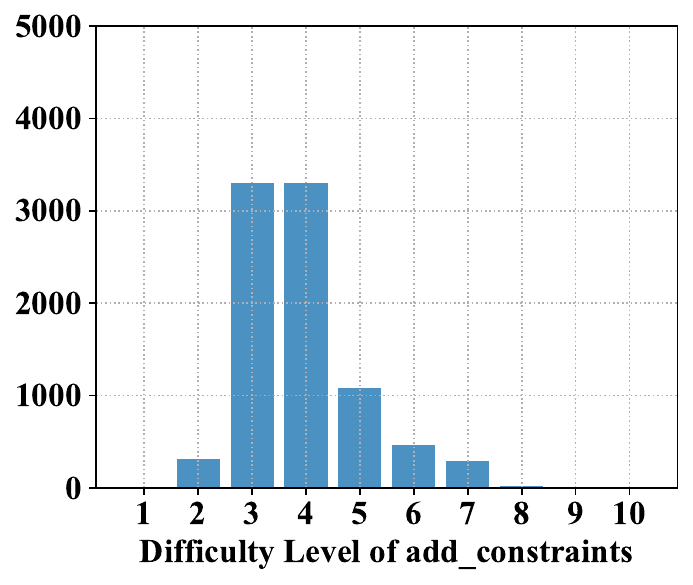}
  \end{subfigure}
  \begin{subfigure}{0.16\linewidth}
    \includegraphics[width=\textwidth]{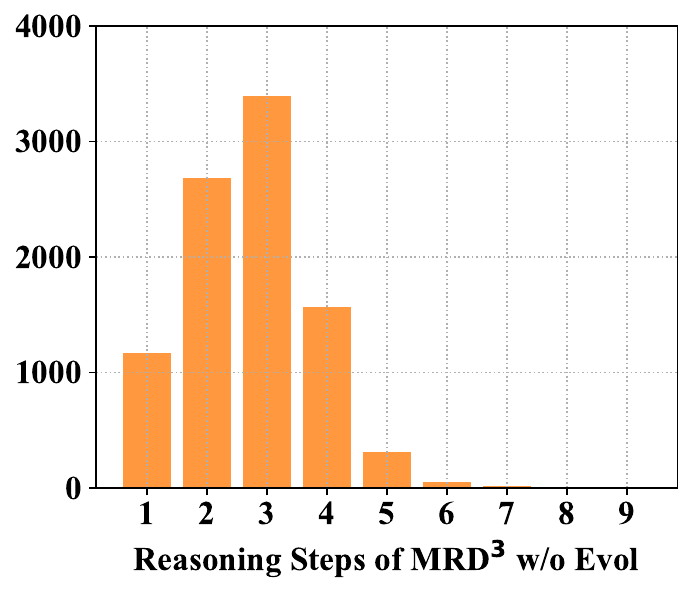}
  \end{subfigure}
  \begin{subfigure}{0.16\linewidth}
    \includegraphics[width=\textwidth]{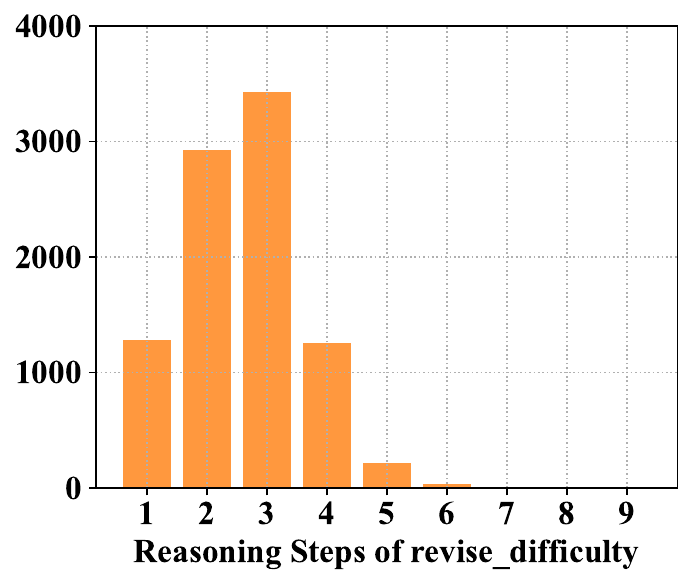}
  \end{subfigure}
  \begin{subfigure}{0.16\linewidth}
    \includegraphics[width=\textwidth]{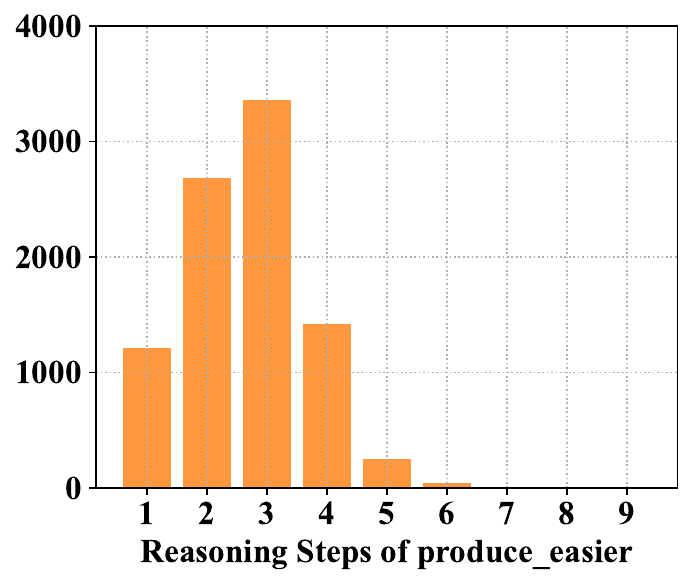}
  \end{subfigure}
  \begin{subfigure}{0.16\linewidth}
    \includegraphics[width=\textwidth]{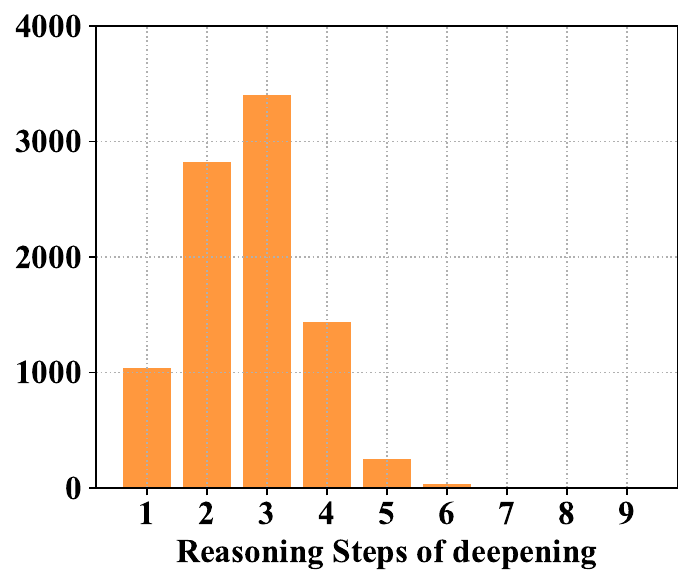}
  \end{subfigure}
  \begin{subfigure}{0.16\linewidth}
    \includegraphics[width=\textwidth]{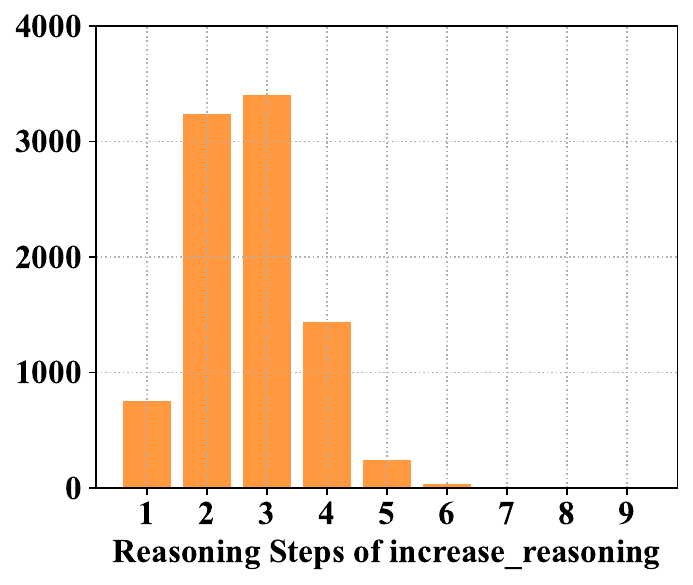}
  \end{subfigure}
  \begin{subfigure}{0.16\linewidth}
    \includegraphics[width=\textwidth]{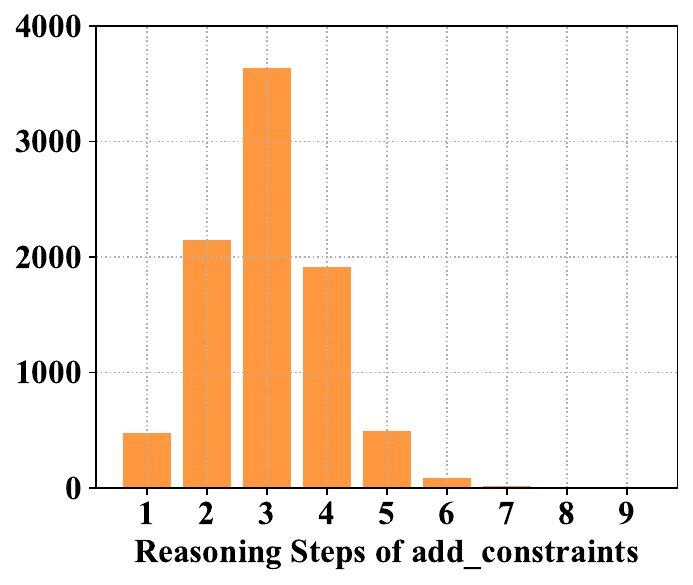}
  \end{subfigure}
  \caption{The difficulty distribution (first row) and the number of reasoning steps distribution (second row). }
  \label{fig:distribution}
\end{figure}

Based on the GPT-4-based estimation, we are able to quantitatively look into the distribution of difficulty and reasoning step distribution in MRD$^3$ without evolution and MRD$^3$ with various evolution schemes. The results are shown in Figure~\ref{fig:distribution}. The original distribution of both difficulty level and reasoning steps of questions centralized between 2 to 4. More questions with higher difficulty using add\_constraints, deepening, and increase\_reasoning. As we discuss in the reward design of our RL pruner, easy questions are important for the stabilization of RL and can help effectively identify the quality of pruned prompt, more easier questions are generated with revise\_difficulty and produce\_easier evolution scheme.

\subsection{Additional observation on difficulty distribution}

As shown in Figure~\ref{fig:distribution}, the difficulty diversity of examples in MRD$^3$ are improved after prompt evolution. We then research into the difficulty distribution of the input examples for in-context learning. The observation is shown as follow in addition to the 3 main observations proposed in Sec.~\ref{sec:pilot_study}:

\textbf{Observation 4}: \textit{LLMs with different capabilities prefer CoT examples of varying difficulties}.

In our further exploration of the optimal selection of CoT examples for improve mathematical reasoning, we observe that LLMs of different capabilities exhibit preferences for CoT examples of varying difficulty levels. As Table~\ref{tbl:cotdifficulty} shows, we categorize each CoT example in the {MRD$^3$-Evol} dataset by difficulty level. We then select the top 16 CoT examples from different groups as few-shot examples for LLaMA2 models. Results show LLaMA2-7b prefers CoT examples with a difficulty level of 3-4, while LLaMA2-13b, more capable, prefers those with a difficulty level of 4 or above. This aligns with intuition: for instance, when assisting a middle school student with math problems, it is more beneficial to provide examples of  moderate difficulty that they can comprehend, whereas for a high school student, examples  with a higher level of difficulty are more useful. 

\begin{table}[h]
	\caption{Smaller, less capable LLMs favor simpler CoT examples, while larger ones prefer more complex ones.}
	\label{tbl:cotdifficulty}
	\centering
 \resizebox{0.6\textwidth}{!}{
	\begin{tabular}{cccc}
		\toprule
		Model&Difficulty ($\leq$ 3) & Difficulty (3-4)&Difficulty ($\geq$ 4)\\
		\midrule
		LLaMA2-7B&14.49&\textbf{15.39}&14.86\\
		LLaMA2-13B&23.81&25.32&\textbf{25.47}\\
		\bottomrule
	\end{tabular}}
\end{table}

In our evaluation of {\sysname} with various LLMs, we found that the shot selection results are consistent with our observation. The average difficulty score and number of reasoning steps for the examples selected by LLaMA2-70B pruner are 3.57 and 3.04, which are higher than the results of LLaMA2-13B are 3.51 and 2.98. The empirical results further support our assumption that LLMs with larger size prefers harder examples than smaller-scale LLMs.

\subsection{Evolution example}

We give an example of a math questions and then show the evolved results of the questions and answers. The evolved results follow our instruction given in Sec.~\ref{sec:evol-prompt}. As can be seen from the evolution results, the answer of input questions can be modified (\textit{e.g.} ground truth answer change from 16 to 12 in add\_constraints). The whole background of the questions can also be replaced (\textit{e.g.} computation target of question change from current tickets at the arcade to final points of a game in produce\_easier). These modification and variation improve the diversity of our prompt candidate sets, which are the building block for the training and reasoning with {\sysname}.

\begin{center}
\small
\begin{tcolorbox}[width=0.98\textwidth,title = {\textbf{Example of prompt evolution}}]
\textbf{Input example for evolution:} \\
Q: At the arcade Dave won 11 tickets. If he spent 5 tickets on a beanie and later won 10 more tickets, how many would he have? \\
A: Dave initially had 11 tickets, spent 5 on a beanie, and later won 10 more, resulting in a total of: 11 - 5 + 10 = 16 tickets. The answer is 16. 
\tcblower
\textbf{Evolution result of add\_constraints:} \\
Q: At the arcade, Dave won 11 tickets. If he spent 5 tickets on a beanie, won 10 more tickets, and gave 4 tickets to his friend, how many would he have? \\
A: Dave initially had 11 tickets, spent 5 on a beanie, won 10 more, and gave 4 to his friend, resulting in a total of: 11 - 5 + 10 - 4 = 12 tickets. The answer is 12.\\
\textbf{Evolution result of deepening:} \\
Q: At the arcade, Dave triumphed in a game and earned 11 tickets. He redeemed 5 of those tickets for a beanie and subsequently won an additional 10 tickets. What is his final ticket count? \\
A: Dave started with 11 tickets, used 5 to acquire a beanie, and later added 10 more to his collection, resulting in: 11 - 5 + 10 = 16 tickets. The answer is 16.\\
\textbf{Evolution result of increase\_reasoning:} \\
Q: Dave won 11 tickets at the arcade and spent 5 on a beanie. Afterward, he won 10 more tickets. Calculate his final ticket count by first finding the remaining tickets after his purchase and then adding the newly won tickets. \\
A: Dave had 11 tickets, spent 5, leaving him with 6. Then he won 10 more, resulting in: 6 + 10 = 16 tickets. The answer is 16.\\
\textbf{Evolution result of revise\_difficulty:} \\
Q: Dave got 11 tickets at the arcade. He used 5 tickets to get a beanie and then won 10 more tickets. How many tickets does he have now? \\
A: Dave started with 11 tickets, used 5 for a beanie, and won 10 more, giving him: 11 - 5 + 10 = 16 tickets. The answer is 16.\\
\textbf{Evolution result of produce\_easier:} \\
Q: In a game, Sarah scored 11 points. She lost 5 points in a penalty and then scored 10 more points. What is her final score? \\
A: Sarah initially scored 11 points, lost 5 due to a penalty, and scored 10 more, resulting in a total of: 11 - 5 + 10 = 16 points. The answer is 16.
\end{tcolorbox}
\end{center}

\section{Pruner Training and Evaluation Details} \label{sec:training_details}

\subsection{Detailed algorithm for training data preparation}
As a supplement to phase 1 in Algorithm~\ref{alg:cot-max}, we show the algorithm for training data preparation in Algorithm~\ref{alg:dataset}. Both the difficulty level and number of reasoning step are involved in the GPT-4-based evaluation. However, we omit the reasoning step in this algorithm as we only use difficulty level in the training set split.

\begin{algorithm}[htb]
\small
   \caption{Training dataset preparation}
   \label{alg:dataset}
   \textbf{Input:} CoT dataset $\{ x^{\text{cot}}_i\}_{i=1}^{L}$, difficulty threshold $d_{\text{thr}}$, \\ 
   \textbf{Output:} MRD$^3$ $\mathcal{D} = \{ x^{\text{cot}}_j, d_j\}_{j=1}^{L^{\text{MRD$^3$}}}$, questions set $\mathcal{D}_{\text{question}}$, prompt set $\mathcal{D}_{\text{cot}}$
\begin{algorithmic}[1]
   \STATE $\blacktriangleright$ \textbf{Phase 1: MRD$^3$-Evol Preparation}
   \STATE MRD$^3$ dataset $\mathcal{D} = \{\}$
   \FOR {$i = 1$ to $L$}
   \STATE Perform GPT-4 based prompt evolution on $x^{\text{cot}}_i$ to get $\{x^{\text{cot-E}}_{i,e}\}_e$
   \STATE Evaluate difficulty on $\{x^{\text{cot-E}}_{i,e}\}_e$ to get score $\{d_{i,e}\}_e$ using GPT-4
   \STATE Append examples $\{x^{\text{cot-E}}_{i,e}, d_{i,e}\}_e$ to $\mathcal{D}$
   \ENDFOR
   \STATE Prompt set $\mathcal{D}_{\text{cot}} = \{\}$, question set $\mathcal{D}_{\text{question}} = \{\}$
   \FOR {$j = 1$ to $L^{\text{MRD$^3$}}$}
   \IF {$d_j \leq d_{\text{thr}}$}
   \STATE Append example $(x^{\text{cot}}_j, d_j)$ to $\mathcal{D}_{\text{question}}$
   \ELSE 
   \STATE Append example $(x^{\text{cot}}_j, d_j)$ to $\mathcal{D}_{\text{cot}}$
   \ENDIF
   \ENDFOR
   \STATE \textbf{Return}  full dataset with evolution $\mathcal{D}$, questions set $\mathcal{D}_{\text{question}}$, prompt candidate set $\mathcal{D}_{\text{cot}}$
\end{algorithmic}
\end{algorithm}

\subsection{Detailed settings and hyperparameters}

The detailed hyper-parameters setting of different LLMs' pruner are listed in Table~\ref{tab:hyperpara}. Majority of these hyperparameters are shared across different LLMs. The evolution subset as the prompt candidates for evaluation are determined by searching the performance of math reasoning on 100 random examples.

\begin{table}[h]
\centering
\caption{Detailed hyper-parameters for pruner training scheme of different LLMs.}
\resizebox{0.8\textwidth}{!}{
\begin{tabular}{c|ccc}
\toprule
Model            & LLaMA2-7B & LLaMA2-13B & LLaMA2-70B    \\ 
\midrule
Epoch                & 3            & 3             & 3     \\
Batch Size           & 1            & 1             & 1      \\
Pruner LLM Base      & LLaMA2-13B  & LLaMA2-13B   & LLaMA2-70B  \\
Input Shot           & 40           & 48            & 48  \\
Input Shot (TopK)    & 32           & 32            & 32  \\
Input Shot (Few-shot)& 8            & 16            & 16  \\
Optimizer            & AdamW        & AdamW         & AdamW      \\
Weight Decay         & 1$e^{-2}$    & 1$e^{-2}$     & 1$e^{-2}$     \\
Learning Rate        & 1$e^{-5}$    & 1$e^{-5}$     & 1$e^{-5}$     \\
\midrule
Embedding Extractor  & BERT-Large (cased) & BERT-Large (cased)  & BERT-Large (cased)   \\
Embedding Size       & 1024         & 1024         & 1024   \\
Tokenizer Padding    & 512          & 512          & 512   \\
Difficulty Threshold $d_{\text{thr}}$   & 2         & 2        & 2 \\
Token Target $T$     & 2048         & 2048         & 2048 \\
Token Penalty Coefficient $w$ & (-1,1)  & (-1,1)   & (-1,1)   \\
Selection Repeat $t_{\text{repeat}}$ & 10 & 10 & 5 \\
\midrule
Evol Subset   & add\_constraints & increase\_reasoning & increase\_reasoning   \\
temperature         & 0.8 & 0.8 & 0.8\\
top\_p              & 0.95 & 0.95 & 0.95\\
top\_k              & 40 & 40 & 40\\
num\_beams          & 1 & 1 & 1\\
max\_new\_tokens    & 1 & 1 & 1\\
\bottomrule
\end{tabular}}
\label{tab:hyperpara}
\end{table}

\subsection{Training dynamics}
\begin{figure}[h]
  \centering
  \begin{subfigure}{0.24\linewidth}
    \includegraphics[width=\textwidth]{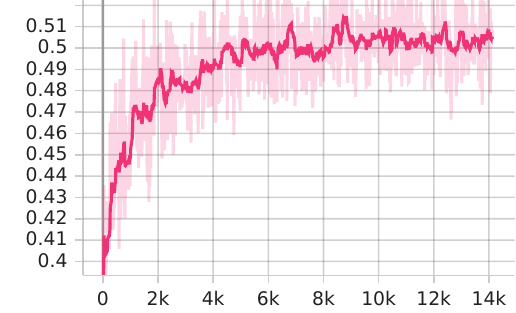}
    \caption{LLM loss reward $\frac{1}{1+L_\text{LLM}}$}
  \end{subfigure}
  \begin{subfigure}{0.24\linewidth}
    \includegraphics[width=\textwidth]{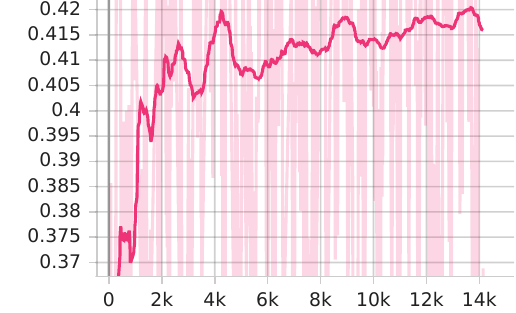}
    \caption{Prediction reward $R_{\text{Acc}}$}
  \end{subfigure}
  \begin{subfigure}{0.24\linewidth}
    \includegraphics[width=\textwidth]{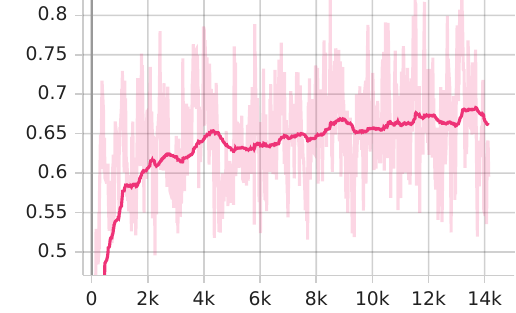}
    \caption{EMA pruner reward $R$}
  \end{subfigure}
  \begin{subfigure}{0.24\linewidth}
    \includegraphics[width=\textwidth]{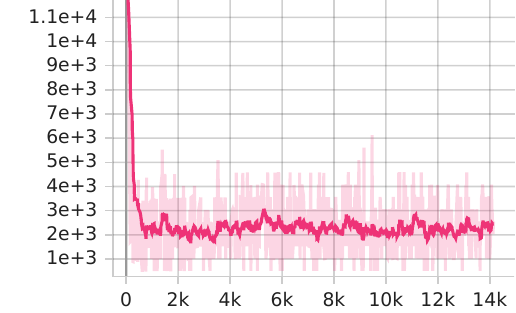}
    \caption{Remaining token $t$}
  \end{subfigure}
  \caption{RL training dynamics of the LLaMA2-13B pruner. }
  \label{fig:reward_curve}
\end{figure}

We visualize the RL training dynamics of the LLaMA2-13B pruner in Figure~\ref{fig:reward_curve} including the LLM loss reward $\frac{1}{1+L_\text{LLM}}$, prediction reward $R_{\text{Acc}}$, moving average of the final pruner reward $R$, and remaining token count $t$. We can see from the results that reward increases steadily with the steps of RL training. The number of remaining tokens decreases rapidly in early steps and then oscillates around the token target. Since our prediction reward $R_{\text{Acc}}$ are discrete value of $\{-0.1, 0, 1\}$, the oscillation phenomenon are more obvious compared with other reward term. This highlight the effectiveness of question repetition and using Exponential Moving Average (EMA) of final reward to suppress this oscillation.

\subsection{Detailed introduction of dataset for evaluation}

We introduce the details of the datasets we used for evaluation as follows:
\begin{itemize}
    \item \textbf{GSM8K}~\cite{gsm8k} is a math reasoning dataset consisting high quality linguistically diverse grade school math word problems created by human problem writers. There are 7473 training examples and 1319 validation examples in the dataset.
    \item \textbf{SVAMP}~\cite{svamp} representing \textbf{S}imple \textbf{V}ariations on Arithmetic Math word Problems that conduct question sensitivity variation, reasoning ability variation, and structural variation on existing math datasets. There is a total of 1000 examples and all of them are used for evaluation in our settings.
    \item \textbf{MultiArith}~\cite{multiarith} is a collection of multi-step arithmetic problems with 600 examples and all of them are used for evaluation in our settings.
    \item \textbf{AddSub}~\cite{addsub} is a dataset consisting of addition and subtraction problems with 395 examples and all of them are used for evaluation in our settings.
    \item \textbf{SingleEq}~\cite{singleeq} consists grade-school algebra word problems that map to single equations with varying length. Every equation may involve multiple math operations including multiplication, division, subtraction, and addition over non-negative rational numbers and only one variable. There are 508 problems, 1117 sentences, and 15292 words in the dataset.
\end{itemize}

\subsection{Rule-based prompt reconstruction}

To make sure the input prompt for inference remain structurally intact, we apply a rule-based prompt reconstruction on the input. For example, ``\textit{$\backslash$n [question]}'' will be reconstructed to ``\textit{$\backslash$nQ: [question]}'' and ``\textit{A: Let's step by step}'' will be repaired to ``\textit{A: Let's think step by step}''. While our pruner has been trained to learn the importance of the structure integrity and consistency, there are still few cases when important tokens are pruned, leading to incorrect reasoning results. The rule-based reconstruction can effectively alleviate the influence of sub-optimal pruning strategy. The important tokens that should be reconstructed include `Q:', `A:', `$\backslash$n', ``Let's think step by step'', and ``The answer is''.

\section{Additional Related Works} \label{sec:baseline}
\noindent\textbf{LLM In-Context Learning} In-context learning (ICL) are one of the emerging abilities of LLMs that conduct various downstream tasks with provided few-shot demonstrations. To fully understand optimize the ICL paradigm, previous research mainly focus on the underlying mechanism of ICL or the proper application of ICL. Pioneering research~\cite{von2023transformers, dai2023can} empirically find the similarity between gradient-descent (GD) and ICL, which interprets the trained LLMs are meta-optimizer that can learn the examples in context in forward pass. More recently, \citet{wang2023label} propose  a hypothesis that label words in examples serve as anchors in ICL, and the anchors can help aggregate and distribute the task-relevant information flow. To better utilize ICL, previous research also research on the input format~\cite{yoo2022ground} and order of examples~\cite{min2022rethinking}. Our work falls in the second category that shows the compressed examples are an optimal choice for the input of ICL.

\noindent\textbf{LLM Context Window Extension}
 Recently, there has been rising interests in extending the context window of  existing pre-trained LLMs. Common approaches include augmenting external memory modules~\cite{longLLaMA,wang2023augmenting}, which add extra modules to memorize long past contexts but requires complex training, manipulating attention mechanisms~\cite{han2023lminfinite, streamingllm} or the positional encoding~\cite{pi,yarn}. However, these require LLM modifications. Our method, applicable to black-box LLMs and extendable context windows, is orthogonal to this direction.

\noindent\textbf{Comparison of {\sysname} with Previous Methods} We summarize the advantage of our {\sysname} compared with previous prompting strategies in Table~\ref{tab:advantage_compare}. \textit{Gradient-free} indicates the methods do not need to backward through LLMs. \textit{Unlimited-token} represents the original input prompt for these methods are not limited by the context window length of LLMs. \textit{Difficulty-aware} refers to whether the method take the difficulty of problems into the consideration of their prompt design. \textit{Dynamic \#Shots} means we do not need to setup a target shot number and the pruned input shot numbers are different across various questions. Our {\sysname} demonstrate significant advantage over all previous methods.

\begin{table}[h]
\caption{Comparison of the advantage of different prompting strategies. }
\label{tab:advantage_compare}
\centering
\resizebox{\textwidth}{!}{
\begin{tabular}{l c c c c c c c c}
\toprule
Methods & Frozen LLMs & Automated & Gradient-free & Unlimited-token & Transferable & Interpretable & Difficulty-aware & Dynamic \#Shots\\
\midrule
Fine-Tuning                           & \XSolidBrush  & \Checkmark  & \XSolidBrush  & \XSolidBrush  & \XSolidBrush & \XSolidBrush & \XSolidBrush & \XSolidBrush \\
Manual Prompt                         & \Checkmark  &  \XSolidBrush  &  \Checkmark  &  \XSolidBrush & \Checkmark & \Checkmark & \XSolidBrush & \XSolidBrush \\
Soft Prompt Tuning                    & \Checkmark  & \Checkmark & \XSolidBrush & \XSolidBrush &  \XSolidBrush & \XSolidBrush & \XSolidBrush & \XSolidBrush \\ 
Prompt Retrieval                      & \Checkmark  & \Checkmark  & \Checkmark & \XSolidBrush  & \Checkmark & \Checkmark &  \XSolidBrush & \XSolidBrush \\
AutoPrompt~\cite{shin2020autoprompt}  & \Checkmark &  \Checkmark & \XSolidBrush & \XSolidBrush & \Checkmark & \Checkmark & \XSolidBrush & \XSolidBrush  \\  
RLPrompt~\cite{rlprompt}              & \Checkmark  & \Checkmark & \Checkmark & \XSolidBrush  & \Checkmark & \Checkmark &  \XSolidBrush & \XSolidBrush \\
Context Extension                     & \Checkmark  & \Checkmark  & \Checkmark  & \Checkmark  & \Checkmark & \Checkmark &  \XSolidBrush & \XSolidBrush \\
LLMLingua~\cite{llmlingua}            & \Checkmark  & \Checkmark  & \Checkmark  & \Checkmark  & \Checkmark & \Checkmark &  \XSolidBrush & \XSolidBrush \\
\midrule
 \sysname (Ours)                      & \Checkmark  & \Checkmark  & \Checkmark  & \Checkmark  & \Checkmark & \Checkmark &  \Checkmark & \Checkmark \\
\bottomrule
\end{tabular}}
\end{table}

\section{Prompt Settings} \label{sec:prompt_setting}

In this section, we show the prompt we used in this work for reproducibility. The prompt for evaluating the difficulty and reasoning steps of each examples are:

\begin{center}
\small
\begin{tcolorbox}[width=0.98\textwidth,title={\textbf{Prompt for difficulty and reasoning steps estimation:} }]

We would like you to evaluate and rate the difficulty and complexity of the following question. You should first give an overall score on a scale of 1 to 10, where a higher score indicates higher difficulty and complexity. You should then evaluate the answer and give how many reasoning steps are in the answer. You must just give the score and the number of reasoning steps without any other reasons. The reply format should be 'Score': [score], 'Steps: [\#steps]' \\
\#\# Question: \textbf{\{\underline{Given question}\}}\\
\#\# Answer: \textbf{\{\underline{Given answer}\}}\\
\#\# Evaluation:
\end{tcolorbox}
\end{center}

The prompt for GPT-4 Compression on prompts are shown as follow. Note that we encode the restriction of token limits in both the prompt and API by setting the \textit{max\_new\_token}. However, the prompt compression results still fail to follow the restrictions for most cases. This disadvantages of uncontrollable token length is also discussed in previous work~\cite{llmlingua}.

\begin{center}
\small
\begin{tcolorbox}[width=0.98\textwidth,title={\textbf{Prompt for GPT-4-based compression:}}] 

Please compress the following examplars for few-shot in-context learning on math reasoning. The complete examplars could be removed if they are redundant and the tokens within each examplars can also be pruned. 'The answer is' in each examplar should be retained and please keep less than \textbf{\{\underline{Given token}\}} tokens in total: \\
\textbf{\{\underline{Given examplars}\}}
\end{tcolorbox}
\end{center}
\end{document}